\documentclass{article}
\usepackage[utf8]{inputenc}
\usepackage[T1]{fontenc}

\usepackage{PRIMEarxiv}

\usepackage[utf8]{inputenc} 
\usepackage[T1]{fontenc}    
\usepackage{url}            
\usepackage{booktabs}       
\usepackage{amsfonts}       
\usepackage{nicefrac}       
\usepackage{enumitem}
\usepackage{multirow}
\usepackage{multicol}
\usepackage{listings}
\usepackage{xcolor}
\usepackage{amsmath}
\usepackage{microtype}      
\usepackage{lipsum}
\usepackage{fancyhdr}       
\usepackage{graphicx}       
\graphicspath{{media/}}     
\usepackage[backend=biber]{biblatex}
\usepackage{hyperref}

\title{INSIGHT: An Interpretable Neural Vision-Language Framework for Reasoning of Generative Artifacts
}

\lstdefinestyle{PromptStyle}{
    basicstyle=\ttfamily\small, 
    breaklines=true,
    columns=fullflexible,
    frame=single, 
    frameround=r, 
    backgroundcolor=\color{gray!10}, 
    captionpos=b,
    numbers=none,
    rulecolor=\color{black}
}

\author{
  Anshul Bagaria \\
  Department of Data Science and Artifcial Intelligence\\
  Indian Institute of Technology, Madras, Tamil Nadu, India\\
  \texttt{anshulr2010@gmail.com}
}

\begin{document}
\maketitle

\begin{abstract}
The growing realism of AI-generated images produced by recent GAN and Diffusion Models has intensified concerns over the reliability of visual media. Yet, despite notable progress in deepfake detection, current forensic systems degrade sharply under real-world conditions such as severe downsampling, compression, and cross-domain distribution shifts. Moreover, most detectors operate as opaque classifiers, offering little insight into why an image is flagged as synthetic—undermining trust and hindering adoption in high-stakes settings.

We introduce \textbf{INSIGHT} (Interpretable Neural Semantic and Image-based Generative-forensic Hallucination Tracing), a unified multimodal framework for robust detection and transparent explanation of AI-generated images, even at \textbf{extremely low resolutions} (16$\times$16 - 64$\times$64). INSIGHT combines (i) \textbf{hierarchical super-resolution} for amplifying subtle forensic cues without inducing misleading artifacts, (ii) \textbf{Grad-CAM–driven multi-scale localization} to reveal spatial regions indicative of generative patterns, and (iii) \textbf{CLIP-guided semantic alignment} to map visual anomalies to human-interpretable descriptors. A vision-language model is then prompted using a structured \textbf{ReAct + Chain-of-Thought} protocol to produce consistent, fine-grained explanations, verified through a dual-stage \textbf{G-Eval + LLM-as-a-judge} pipeline to minimize hallucinations and ensure factuality.

Across diverse domains—including animals, vehicles, and abstract synthetic scenes, INSIGHT substantially improves both detection robustness and explanation quality under extreme degradation, outperforming prior detectors and black-box VLM baselines. Our results highlight a practical path toward transparent, reliable AI-generated image forensics and establish INSIGHT as a step forward in trustworthy multimodal content verification.
\end{abstract}


\section{Introduction}
In the last decade, generative models have evolved
from producing coarse, artifact-ridden images to
synthesizing visuals of astonishing realism. Alongside rapid progress in image generation, the modern landscape of generative AI is increasingly
\textbf{multimodal}, with models now capable of synthesizing not only images but also coherent audio tracks, video sequences, and other sensory modalities. Recent systems demonstrate \textbf{unified architectures} that operate across vision, audio, and 
motion domains (\autocite{a_liu2024sora}, \autocite{a_lu2024unified}, \autocite{a_maaz2024video}, \autocite{a_zhang2024mm}), 
signaling a shift from single-task generators toward general-purpose media synthesis engines. As these models continue to scale and absorb ever-larger multimodal corpora, the perceptual gap between human-authored and machine-generated media narrows at an accelerating pace.

The trajectory of generative modeling reflects this trend. Early GAN-based approaches
\autocite{1_goodfellow2014generativeadversarialnetworks, 1_arjovsky2017wassersteingan}
introduced the first wave of AI-synthesized visuals—promising but limited, often displaying texture inconsistencies or structural distortions. In contrast, modern diffusion-based architectures
\autocite{ho2020denoisingdiffusionprobabilisticmodels, rombach2022highresolutionimagesynthesislatent}
can generate high-resolution, \textbf{semantically consistent images} conditioned on free-form natural language prompts. Their utility spans diverse, media-intensive domains: generating synthetic medical imagery for data augmentation \autocite{3_karras2019stylebasedgeneratorarchitecturegenerative}, powering virtual try-on systems in fashion technology \autocite{4_han2018vitonimagebasedvirtualtryon}, designing video game environments, and enabling professional-grade digital art pipelines.

This rapid evolution has democratized content creation, enabling artists, developers, and even non-expert users to produce photorealistic imagery in seconds. At the same time, the increasing realism and multimodality of generated content create new challenges for digital forensics: distinguishing authentic media from synthetic counterparts becomes progressively more difficult, especially in bandwidth-constrained or degraded settings common to social media and messaging platforms. Essentially, this widespread use of AI-generated images has also raised serious concerns about visual reliability, semantic plausibility, and interpretability.

\subsection{\textbf{The Stakes Are No Longer Limited to Deepfake Faces}}
The stakes of synthetic media generation now extend far beyond the early discourse around deepfake faces. Although public concern was initially centered on fabricated human identities and manipulated political videos \autocite{b_yadav2019deepfake}, \autocite{b_agarwal2020detecting}, contemporary generative models have dramatically broadened the landscape of what can be realistically forged. Systems such as \textbf{Stable Diffusion} \autocite{rombach2022highresolutionimagesynthesislatent}, \textbf{Midjourney} \autocite{b_jaruga2022artificial}, and \textbf{DALL-E} \autocite{dall-e} routinely produce hyperrealistic wildlife photographs, autonomous-driving street scenes, medical radiology scans, and even satellite-style geospatial renderings indistinguishable from operational surveillance data (\autocite{c_zhou2025freeblend}, \autocite{c_koetzier2024generating}, \autocite{c_mai2025towards}). As a result, domains once assumed to be naturally resistant to AI-driven manipulation - ecology, transportation, clinical diagnostics, environmental monitoring, and defense intelligence - are increasingly vulnerable to high-fidelity synthetic fabrication \autocite{chamot2022deepfake}.

This expansion fundamentally alters the forensic threat model. The generative capabilities affecting these domains introduce \textbf{risks} that are not merely aesthetic or reputational but operational, influencing high-stakes decision-making pipelines. For instance, synthetic wildlife imagery can distort ecological datasets, fabricated traffic scenes may compromise safety validation in autonomous driving, and artificially generated medical images pose critical hazards to diagnostic workflows and dataset curation. Similarly, forged satellite or aerial imagery can mislead humanitarian assessments or escalate geopolitical tensions.

Consequently, the threat surface is no longer narrow or domain-specific, \textbf{it is universal}, spanning any field that relies on visual evidence or image-based analytics. This universality heightens the urgency for forensic techniques that can not only distinguish synthetic media from real imagery but also explain the reasoning behind such determinations, especially in degraded or low-resolution settings typical of online dissemination platforms \autocite{kumar2025ai}. It is in this context that interpretability, low-resolution robustness, and cross-domain generalization become essential pillars for the next generation of AI-forensic methodologies.

\subsection{\textbf{The Fragility of AI-Generated Image Detectors in Practical Settings}}
Despite remarkable progress in synthetic image detection, existing forensic methods are designed under laboratory conditions. As generative models permeate an ever-wider range of domains, the corresponding forensic burden shifts from controlled academic benchmarks to the fragmented, unpredictable conditions of real-world media circulation. In practice, synthetic images are rarely encountered in their original resolution; instead, they propagate through social networks, messaging platforms, automated content pipelines, and low-bandwidth communication channels—each introducing its own combination of \textbf{downsampling}, \textbf{recompression}, \textbf{color subsampling}, and \textbf{algorithmic filtering}. These transformations systematically degrade or eliminate the subtle forensic cues that detectors rely on. As a result, the challenge is no longer simply to identify synthetic signals, but to do so under extreme visual degradation and significant distribution shift, where the available evidence is sparse, quantized, or severely distorted. This mismatch between laboratory assumptions and deployment realities underscores the need to examine the unique difficulties posed by low-quality, heavily processed imagery. This hereby motivates the design of forensic systems explicitly tailored for such conditions.

A major consequence of this real-world degradation pipeline is the weakening, or complete disappearance, of visual artifacts traditionally used to flag synthetic content. Such artifacts, which arise from generator weaknesses or training biases, can include subtle texture irregularities or more overt semantic implausibilities, such as duplicated fingers, malformed vehicle components, or inconsistent symmetries \autocite{6_Gragnaniello2022}, \autocite{7_marra2018gansleaveartificialfingerprints}. While detectors often depend on these artifacts as discriminative cues, their reliability decays sharply when images undergo compression, denoising, or aggressive scaling. In low-resolution scenarios, where only coarse structural information survives, high-frequency traces are smeared or lost entirely, rendering artifact-based detection both technically fragile and visually ambiguous \autocite{8_wang2018esrganenhancedsuperresolutiongenerative}.

\subsection{\textbf{Beyond Accuracy: The Crisis of Meaningful Explanations}}
While robustness under real-world degradation is a fundamental challenge, an equally pressing, and often under-examined, problem concerns the interpretability of forensic systems. Even when state-of-the-art detectors succeed in identifying synthetic media, they rarely articulate why a specific region is suspicious. Popular post-hoc explainability tools such as \textbf{Grad-CAM} \autocite{Selvaraju_2019}, and \textbf{gradient-based saliency maps} \autocite{simonyan2014deepinsideconvolutionalnetworks} provide only coarse heatmaps that highlight spatial regions of interest without offering semantic clarity about the underlying generative flaws. As a result, investigators are left with visually diffuse cues that do not meaningfully distinguish between trivial correlations and genuine forensic evidence.

Recent efforts have explored the use of \textbf{vision–language models} (VLMs) as interpretable explainers for synthetic imagery \autocite{yin2025digitalforensicsagelarge}, \autocite{li2022blipbootstrappinglanguageimagepretraining}. While promising, these systems introduce their own failure modes. VLMs often fabricate causal narratives, \textbf{hallucinate} nonexistent artifacts, or misinterpret generative patterns when visual evidence is weak or noisy—a common occurrence in compressed or low-resolution environments. Such hallucinated explanations can undermine forensic reliability and create false confidence in downstream decisions.

Prior research in image forensics, deepfake detection, and authenticity verification has largely focused on \textbf{classification accuracy} \autocite{9_dolhansky2020deepfakedetectionchallengedfdc}, \autocite{10_Verdoliva_2020}. Many systems behave as black boxes, optimizing for binary real-vs-fake predictions while offering little insight into the specific visual cues that motivated their decisions. Other approaches rely on noise residuals, fingerprints, or \textbf{frequency-domain inconsistencies} \autocite{11_qi2025sfnetfusionspatialfrequencydomain}. Although these signals can be powerful, they often fail to generalize across generator families, architectural shifts, or diffusion-based pipelines, thereby limiting their applicability as universal forensic tools. More importantly, such low-level cues seldom translate into \textbf{human-understandable explanations} that can assist auditors, policymakers, or stakeholders in evaluating the trustworthiness of an image.

This absence of transparency presents a broader crisis. Without semantically grounded explanations for detected artifacts, forensic outputs lack accountability and are difficult to validate in high-stakes domains such as healthcare diagnostics, scientific publishing, autonomous driving, or legal evidence analysis \autocite{12_brundage2020trustworthyaidevelopmentmechanisms}, \autocite{13_kuhn2025efficientunsupervisedshortcutlearning}. Misidentified or hallucinated artifacts can mislead practitioners, produce erroneous narratives, or be exploited in adversarial scenarios. As deep generative models continue to accelerate in quality and diversity, the gap between binary detection and trustworthy interpretation widens. Without trustworthy explanations, forensic results are difficult for investigators, journalists, or legal institutions to rely upon. \textbf{Opacity breeds skepticism}. The central question - \textit{what can be believed?} - no longer pertains only to authenticity but also to the explainability of the forensic tools themselves.

\subsection{\textbf{The Core Gap: Absence of a Unified, Trustworthy Pipeline for Low-Resolution Forensics}}
Despite rapid progress across image enhancement, synthetic-media detection, and explainability techniques, the field lacks an integrated system capable of delivering reliable and interpretable forensics under extreme low-resolution conditions. Existing research efforts typically address only \textbf{isolated components} of the pipeline. Super-resolution models attempt to restore detail but frequently introduce hallucinated textures or synthetic edges that contaminate the forensic evidence \autocite{cao2022referencebasedimagesuperresolutiondeformable}. Authenticity classifiers focus on binary real/fake predictions but are not optimized to operate on heavily compressed or low-information inputs. Meanwhile, interpretability approaches—including \textbf{CLIP-based vision–language systems} \autocite{radford2021learningtransferablevisualmodels} and LLM-based explainers often misattribute semantic meaning to noise, overfit to prompt structure, or produce explanations that cannot be visually verified.

These fragmented solutions are not merely incomplete, they are fundamentally misaligned with the needs of real-world forensic analysis. Enhancement models prioritize perceptual quality rather than authenticity. Detectors prioritize accuracy rather than semantic grounding. Explainability tools highlight spatial regions without confirming whether the content corresponds to genuine generative artifacts. When deployed in isolation, each component risks amplifying errors from the others, especially in low-resolution cases where signal strength is already marginal.

What is conspicuously missing is a \textbf{cohesive multimodal framework} that simultaneously \textbf{(i)} strengthens weak forensic cues without fabricating misleading artifacts, \textbf{(ii)} identifies and localizes synthetic patterns robustly across degradations, and \textbf{(iii)} produces explanations that are semantically grounded, visually verifiable, and resilient to hallucinations.
    
A truly reliable system must not treat enhancement, detection, and explanation as disjoint modules but rather as interdependent processes whose outputs must remain consistent and mutually reinforcing. The absence of such a unified, integrity-preserving pipeline represents a critical gap in current forensic capabilities—and directly motivates the framework we propose.

\subsection{\textbf{Our Work: INSIGHT}}
To address these challenges, we introduce \textit{INSIGHT} (Interpretable Neural Semantic and Image-based Generative-forensic Hallucination Tracing), a multimodal forensic system designed from the ground up to operate in the harshest conditions—tiny, noisy, compressed images where both humans and detectors typically fail. INSIGHT \footnote{https://github.com/anshul-2010/INSIGHT} contributes four pillars toward trustworthy low-resolution forensics:
\begin{itemize}
    \item \textbf{Hierarchical Forensic Super-Resolution}: A signal-aware SR model amplifies subtle generative patterns without injecting misleading hallucinations, enabling downstream models to detect cues that would otherwise be invisible.
    \item \textbf{Artifact Localization Through Multi-Scale Attention Fusion}: We integrate Grad-CAM, cross-layer attention, and multi-resolution feature attribution via super-pixelation to pinpoint exactly where synthetic artifacts emerge in reconstructed images.
    \item \textbf{Semantic Grounding with CLIP-Based Forensic Alignment}: Localized artifacts are matched to interpretable descriptors (e.g., “unnatural texture repetition,” “implausible edge transitions”), bridging the gap between pixel-level anomalies and human-language reasoning.
    \item \textbf{Validated Natural-Language Explanations via ReAct + Chain-of-Thought}: A vision-language LLM interprets visual cues through a structured reasoning protocol, with outputs verified through G-Eval [Liu et al. 2023] and a separate VLM judge to minimize hallucination and ensure factual consistency.
\end{itemize}
INSIGHT significantly improves detection robustness under extreme degradation across diverse domains, including animals, vehicles, natural scenes, and abstract synthetic imagery, and produces detailed explanations that both experts and lay users can trust. By tightly coupling measurable forensic signals with validated semantic reasoning, INSIGHT sets a new direction for transparent, reliable, and domain-general AI-generated image forensics.

\section{Related Work}
\label{sec:related_work}

\subsection{\textbf{Artifact Detection in AI-Generated Images}}
The detection and interpretation of artifacts in AI-generated imagery has evolved significantly alongside advances in generative modeling. Early work on GAN forensics established that generative models leave behind subtle, model-dependent \textbf{fingerprints} in local frequency bands, interpolation kernels, and upsampling operations (\autocite{7_marra2018gansleaveartificialfingerprints}, \autocite{ab_1_yu2019attributingfakeimagesgans}, \autocite{ab_2_li2018detection}, \autocite{ab_3_nguyen2024laa})). These artifacts arise because GANs depend on \textbf{learned upsampling filters} and \textbf{noise injection processes} that systematically differ from natural image statistics. Li et al. (2018) \autocite{ab_2_li2018detection} showed that GAN upsampling produces checkerboard artifacts, while Zhang et al. (2019) \autocite{21_zhang2019detectingsimulatingartifactsgan} and Marra et al. (2019) \autocite{7_marra2018gansleaveartificialfingerprints} identified repetitive texture and abnormal frequency distributions. Subsequent studies showed that even minor changes in a generative pipeline can alter its fingerprint, motivating techniques that isolate intrinsic local cues rather than relying solely on global image structure. Subsequent analyses of CNN behavior further reinforced this view: Geirhos et al. (2019) \autocite{ab_4_geirhos2022imagenettrainedcnnsbiasedtexture} demonstrated that neural networks often rely heavily on localized texture patches, and that such micro-level differences play a crucial role in classification outcomes. 

This emphasis on local signals led to the rise of patch-based detectors, where classifiers intentionally operate on restricted receptive fields to avoid global semantics and instead focus on the artifact-bearing regions most indicative of synthesis (\autocite{ab_5_chai2020makesfakeimagesdetectable}, \autocite{ab_6_mahara2025methodstrendsdetectingaigenerated}). More recent studies extend this trend by extracting noise fingerprints or texture-driven forensic traces from a single patch, enabling more robust generalization across diverse GAN and Diffusion architectures. However, most approaches deteriorate under extreme downsampling or compression, where forensic cues become faint or disappear entirely, highlighting the need for forensic-oriented super-resolution and robust multimodal interpretation.

\subsection{\textbf{Super-Resolution for Forensic and Perceptual Enhancement}}
A parallel challenge emerges when considering low-resolution synthetic images, such as those present in \textbf{CIFAKE} (\autocite{ab_8_bird2023cifakeimageclassificationexplainable}), where forensic cues become substantially harder to recover. From an information-theoretic standpoint, deep networks risk progressively discarding subtle high-frequency signals as activations propagate through successive layers (\autocite{ab_9_tishby2015deeplearninginformationbottleneck}). To mitigate this bottleneck, the super-resolution literature offers a spectrum of reconstruction-oriented techniques ranging from interpolation-based (\autocite{ab_10_zhang2018learningsingleconvolutionalsuperresolution}, \autocite{ab_11_qin2020multi}) to learning-based (\autocite{ab_12_dong2015imagesuperresolutionusingdeep}, \autocite{ab_13_tian2024generativeadversarialnetworksimage}, \autocite{ab_14_liu2018attentionbasedapproachsingleimage}). Super-resolution (SR) has traditionally focused on \textbf{perceptual quality} rather than forensic reliability. Early interpolation approaches (\autocite{ab_10_zhang2018learningsingleconvolutionalsuperresolution}, \autocite{223_sun2019super}) provided structural enhancement but often smoothed out forensic signals. The introduction of learning-based SR, beginning with \textbf{SRCNN} (\autocite{ab_12_dong2015imagesuperresolutionusingdeep}), \textbf{ESRGAN} (\autocite{ab_12_wang2018esrganenhancedsuperresolutiongenerative}), and \textbf{SRGAN} (\autocite{ab_12_ledig2017photorealisticsingleimagesuperresolution}), dramatically improved texture realism but raised concerns regarding hallucinated content. Perceptual and adversarial losses, while visually appealing, introduce synthetic textures that can pollute downstream forensic analysis (\autocite{ab_12_lugmayr2020srflowlearningsuperresolutionspace}).

More recent architectures incorporate transformer backbones. Transformer-driven approaches such as \textbf{IPT} (\autocite{ab_15_chen2021pretrainedimageprocessingtransformer}), \textbf{SwinIR} (\autocite{ab_16_liang2021swinir}), and \textbf{UFormer} (\autocite{ab_17_wang2021uformergeneralushapedtransformer}) further push the frontier by exploiting long-range dependencies to improve global coherence. However, they still optimize for perceptual fidelity, not forensic fidelity. For our purposes, controlled SR methods must recover real high-frequency cues while strictly avoiding hallucination.  Particularly relevant to forensic restoration are architectures like \textbf{DRCT} (\autocite{hsu2024drctsavingimagesuperresolution}), which emphasize structural preservation over perceptual hallucination—an essential property when upsampling images for artifact analysis rather than aesthetic enhancement. Nonetheless, no existing system integrates SR explicitly as a \textbf{forensic amplifier}—a gap that motivates \textit{INSIGHT}’s hierarchical super-resolution pipeline designed to reveal artifacts that become invisible in low resolution.

\subsection{\textbf{Superpixelation for Patch-Wise Scoring and Improved Localization}}
A central challenge in artifact-based synthetic image forensics is identifying where in the image discriminative cues actually reside. Most existing forensic pipelines rely on fixed grid patches, typically square crops of uniform size, to extract local signals \autocite{dosovitskiy2020image}. While effective for controlled datasets, such rigid partitioning does not account for the underlying structure of real images. Artifacts generated by modern GAN and Diffusion models often \textbf{do not align with rectangular boundaries}; instead, they follow semantic or textural contours, appearing along edges, uniform surfaces, or regions with repeated patterns. As a result, patch-based systems frequently (i) \textbf{split artifact regions} across patch boundaries, (ii) \textbf{dilute discriminative signals} by mixing artifact pixels with clean background pixels, or (iii) \textbf{focus on irrelevant local regions} due to uniform patching rather than content-aware segmentation.

Superpixel segmentation offers a principled solution to these limitations by decomposing images into perceptually meaningful, texture-consistent regions. Classical algorithms such as \textbf{SLIC} (Simple Linear Iterative Clustering) \autocite{ac_1_achanta2012slic}, \textbf{ERS} (Entropy Rate Segmentation) \autocite{ac_2_liu2011entropy}, \textbf{LSC} (Linear Spectral Clustering) \autocite{ac_3_li2015superpixel}, and \textbf{SEEDS} (Superpixels Extracted via Energy-Driven Sampling) \autocite{ac_4_van2012seeds} cluster pixels into irregularly shaped segments that respect local color statistics, gradients, and spatial continuity. Unlike fixed grid patches, superpixels adhere to object boundaries, reflect homogeneous textures, and naturally group visually similar pixels. This boundary-sensitive partitioning has been widely leveraged in segmentation, computational photography, saliency detection, and perceptual grouping due to its efficiency and strong adherence to human visual organization.

For forensic analysis, these properties resonate directly with the nature of artifacts introduced by generative models. GAN-based upsampling often produces subtle checkerboard patterns, directional textures, or noise residuals concentrated along certain structures; diffusion models may leave grain inconsistencies or blending artifacts at object boundaries. Because superpixels form texture-driven clusters, they inherently \textbf{capture these micro-level anomalies} more faithfully than square patches. Moreover, superpixel-based decomposition maintains region integrity, preventing artifact signals from being diluted across neighboring, semantically unrelated pixels.

Despite these advantages, superpixelation has been under-explored in the context of synthetic image forensics. Existing patch-based methods treat the image as a uniform grid rather than a structured composition of local surfaces. This disconnect becomes even more pronounced in low-resolution settings, such as CIFAKE-like datasets, where artifact regions may occupy only a few contiguous pixels. Superpixel segmentation provides an adaptive, resolution-aware mechanism to retain signal-rich regions for downstream analysis, particularly when combined with attention mechanisms or attribution methods like GradCAM.

In the INSIGHT pipeline, we leverage superpixelation not merely as a pre-processing step but as a core component of multi-scale artifact reasoning. Superpixels provide \textbf{adaptive region proposals}, aligning our patch extraction process with real local texture structures. When combined with GradCAM attribution, superpixels allow the system to isolate and prioritize \textbf{discriminative artifact-bearing regions} with high precision. By applying super-resolution selectively to superpixel segments, the pipeline \textbf{avoids unnecessary hallucination} while amplifying the areas where forensic information is most concentrated. Superpixel regions serve as anchors for \textbf{CLIP-based semantic scoring}, enabling text–image alignment to operate on meaningful, artifact-localized components rather than arbitrarily segmented squares.

This synergy addresses a fundamental gap in the literature: existing methods either focus on global classification or rely on rigid patching, neither of which fully leverage the perceptual organization of images. By incorporating superpixel-guided segmentation, INSIGHT achieves a more faithful and interpretable mapping between generative artifacts, their localized spatial footprints, and the multimodal reasoning required to explain them.

\subsection{\textbf{Explainability via GradCAM and Attention Mechanisms}}
Explainability has emerged as another vital dimension, as binary real-vs-synthetic labels offer little insight into why detections are made. To address this, explainability methods such as \textbf{GradCAM} (\autocite{Selvaraju_2019}), \textbf{Score-CAM} (\autocite{ac_6_wang2020score}), and \textbf{LayerCAM} (\autocite{ac_7_jiang2021layercam}) have become essential tools for understanding how vision models make decisions. \textbf{GradCAM} (\autocite{Selvaraju_2019}), however, remains a cornerstone technique for visual interpretability, generating class-specific saliency maps that highlight regions most influential to a model’s decision. In forensic contexts, such \textbf{gradient-based attention} has been used to identify manipulated regions in deepfakes, adversarial perturbations, and GAN-generated content (\autocite{ab_20_liang2025ferretnetefficientsyntheticimage}, \autocite{ab_21_luan2024interpretable}). Notably, prior work showed that while real images distribute relevance broadly, synthetic images often concentrate discriminative cues into only a few artifact-rich patches, providing strong motivation for our localized attention–guided patch selection strategy.

Bird \& Lotfi (2023) \autocite{CIFAKE} further showed that CNN classifiers trained on low-resolution synthetic datasets (e.g., CIFAKE) disproportionately focus on localized artifact-rich patches rather than broad semantic regions. This behavior underscores the need to incorporate explainability into the forensic pipeline, not merely for post-hoc visualization, but as a mechanism to guide selective patching, enhance SR on artifact regions, and ground semantic descriptors. Existing work, however, stops at interpretation and rarely integrates attention maps into a structured reasoning or multimodal alignment framework, leaving room for systems like INSIGHT to leverage GradCAM not just for explanation but for targeted artifact extraction.

\subsection{\textbf{Vision-Language Alignment with CLIP and Similarity-Based Scoring}}
The rise of Vision–Language Models (VLMs) has drastically expanded the scope of multimodal reasoning in forensic pipelines. Models such as \textbf{CLIP} (\autocite{ab_22_radford2021learningtransferablevisualmodels}) demonstrated the strength of \textbf{contrastive pretraining} in aligning images and text within a unified embedding space, enabling robust zero-shot transfer and semantic anomaly detection. More advanced VLMs, including \textbf{LiT} (\autocite{acc_1_zhai2022lit}), \textbf{OpenFlamingo} (\autocite{acc_2_awadalla2023openflamingo}), and \textbf{MOLMO} (\autocite{ab_23_deitke2024molmopixmoopenweights}), offer improved grounding, cross-modal retrieval, and attribute-level reasoning. Building on this paradigm, MOLMO (\autocite{ab_23_deitke2024molmopixmoopenweights}) and related VLMs (\autocite{ab_24_liu2023visualinstructiontuning}, \autocite{ab_25_zhu2023minigpt4enhancingvisionlanguageunderstanding}, \autocite{ab_26_bao2022vlmounifiedvisionlanguagepretraining}) have achieved impressive performance in fine-grained localization, visual attribute reasoning, and defect detection—capabilities directly relevant to identifying and describing generative artifacts. 

Recent adaptations of VLMs for synthetic image analysis include methods that tune vision-language alignment using low-rank modifiers (\autocite{ab_27_liu2024mixturelowrankexpertstransferable}), leverage complementary image–text signals to infer the source generative model (\autocite{ab_28_keita2025ravidretrievalaugmentedvisualdetection}), or extend CLIP with prompt-tuning strategies for deepfake detection while preserving both visual and textual semantics (\autocite{ab_29_chang2024antifakepromptprompttunedvisionlanguagemodels}). These studies reveal that multimodal models hold substantial promise for explaining why an image is synthetic. However, they often lack (i) patch-level grounding, (ii) explicit localization, and (iii) structured, validated explanations. CLIP is seldom integrated directly with GradCAM-based localization or superpixel-based attribution, a critical gap that INSIGHT bridges by aligning localized forensic cues with descriptive natural-language concepts.

\subsection{\textbf{Language Models for Explanation and Judging}}
The increasing reliance on generative models in critical visual domains has intensified the need not only for \textit{binary forensic decisions} but also for \textbf{transparent}, \textbf{human-understandable explanations}. While early deepfake and generative-image detectors largely returned black-box class labels, modern forensic pipelines demand \textbf{structured reasoning} that articulates why an image appears synthetic, what visual cues support that judgment, and whether the explanation itself is trustworthy. This shift has been enabled by recent advances in Large Language Models (LLMs), which now play a pivotal role in bridging raw visual evidence and interpretable forensic insights.

\subsubsection{\textbf{Chain-of-Thought and Structured Reasoning in LLMs}}
Chain-of-Thought (CoT) prompting introduced by Wei et al. (2022) \autocite{18_wei2023chainofthoughtpromptingelicitsreasoning} represented a major advance in enabling LLMs to produce \textbf{intermediate reasoning steps} rather than direct answers. CoT has been applied extensively in symbolic reasoning, mathematical problem-solving, and commonsense inference, demonstrating that natural language reasoning decompositions significantly improve accuracy and stability. However, despite its wide adoption in NLP, CoT remains underused within visual forensics, where most methods still rely on shallow text-generation for describing artifacts or manipulated regions.

Prior research in LLM-based reasoning (\autocite{b_1_wang2024can}, \autocite{b_2_kojima2022large}, \autocite{b_3_creswell2022selection}) shows that structured verbalization helps prevent hallucination, improves factual grounding, and encourages transparent justification - qualities that are essential for describing generative artifacts that often manifest subtly and ambiguously. Nevertheless, forensic applications require more than reasoning alone: they require reasoning that is conditioned on local visual evidence and validated against the image itself, an area where past work is strikingly sparse.

\subsubsection{\textbf{ReAct: Integrating Reasoning with Tool Use and Visual Grounding}}
The ReAct framework (\autocite{17_yao2023reactsynergizingreasoningacting}) extended CoT by coupling \textbf{reasoning traces with action steps}, enabling LLMs to act as agents that iteratively \textbf{observe}, \textbf{infer}, \textbf{verify}, and \textbf{update} their conclusions. ReAct has been widely adopted in planning tasks, embodied agents, retrieval-augmented LLMs, and interactive reasoning environments.

Despite its strengths, only a handful of works have explored ReAct-like protocols in visual tasks, and almost none have applied it to artifact-level forensic reasoning. Existing visual analysis systems (e.g., VQA, captioning) typically produce linear, unverified descriptions without explicit reasoning steps. In contrast, forensic interpretation requires a structured narrative: \textit{What artifact category should look like (prior knowledge)?}, \textit{What is visually observed in the localized patch?}, \textit{How the observation matches or contradicts typical generative artifacts?}, and a \textit{final, justified conclusion}.

This gap motivates frameworks like INSIGHT, which adapt ReAct to multimodal forensics by connecting CoT reasoning with patch-level visual grounding from CLIP and GradCAM.

\subsubsection{\textbf{Evaluating LLM Explanations: Rubric-Based and Model-Based Judging}}
As LLMs increasingly generate explanations, the need for reliable evaluation mechanisms has become critical. Early work on model interpretability relied on saliency maps and local linear approximations, but these methods often fail on low-information inputs. Recent research introduces LLMs as \textbf{post-hoc natural language explainers}, where the model receives visual evidence and outputs human-readable reasoning. For instance, \textbf{BLIP-2} \autocite{a_li2023blip} connects frozen vision encoders with LLMs to perform flexible question-answering and explanation without task-specific training. Similarly, \textbf{LLaVA} \autocite{20_liu2023visualinstructiontuning} and its successors demonstrate that aligning vision transformers with LLMs yields models capable of coherent explanation, justification, and multi-step reasoning over images.

G-Eval (\autocite{19_liu2023gevalnlgevaluationusing}) emerged as one of the first structured evaluators that score text outputs using rubrics aligned with human annotators. G-Eval demonstrated that L\textbf{LM-based judges} can match or exceed the consistency of crowd workers when provided with explicit evaluation dimensions (\textbf{clarity}, \textbf{coherence}, \textbf{relevance}, etc.). Other works (\autocite{b_1_Hashemi_2024}, \autocite{b_2_kim2023prometheus}) later extended rubric-based LLM evaluation to domains such as summarization, reasoning tasks, and safety audits. However, these evaluators remain primarily text-only, meaning they cannot verify whether an explanation about a \textbf{visual artifact} is substantiated by the actual image. This disconnect is especially problematic in synthetic-media forensics, where hallucinated or incorrect explanations can undermine trust even if the classifier is correct.

Multimodal judging, where the judge jointly considers the image and the explanation, is still an underexplored area. Although \textbf{GPT-4V}, \textbf{LLaVA}, and other VLMs have shown promise in visual consistency checking (e.g., hallucination detection in image captions), they have not been systematically integrated into artifact verification pipelines. Existing work instead focuses on caption reliability, safety compliance, or story grounding, not forensic validation.

Unlike traditional computer vision pipelines, LLM-based vision-language models can process both \textbf{perceptual cues} (color, texture, geometry) and \textbf{semantic priors} (object affordances, physical plausibility). Models such as \textbf{GPT-4V} \autocite{c_1_yang2023dawn}, \textbf{Kosmos-2} \autocite{c_1_peng2023kosmos}, and \textbf{Qwen-VL} \autocite{c_1_wang2024qwen2} demonstrate advanced reasoning capabilities, including \textit{multi-hop chains} of visual inference, \textit{physical reasoning} about occlusion and lighting, \textit{contextual inference} based on world knowledge, and \textit{open-ended hypothesis} generation. For forensic applications, this enables a system to judge whether an upscaled or enhanced image is plausible given real-world constraints, helping detect hallucinations introduced by super-resolution or generative restoration.

Recent work reframes LLMs as \textbf{automatic evaluators} (LLM judges), where the language model assesses the correctness, coherence, or safety of output generated by another model. Papers like \textbf{MT-Bench} and \textbf{Chatbot Arena} \autocite{c_2_zheng2023judging} show that LLM judges can produce human-aligned evaluations with remarkable consistency. Vision-language variants extend this approach to image-based tasks: the model scores reconstructions, flags inconsistencies, or explains deviations from expected semantics.

This judge-based paradigm is particularly relevant for \textbf{low-resolution} image forensics, where traditional \textbf{pixel-level metrics (PSNR/SSIM)} break down. \textit{INSIGHT} directly fills this gap by combine superpixel-based evidence decomposition, resolution-aware enhancement, and LLM-driven reasoning
into a \textbf{single coherent forensic pipeline}. This unified approach closes a significant gap by enabling models to not only enhance degraded images but also justify their outputs, assess evidence consistency, and provide legally and operationally meaningful explanations in forensic contexts.

\section{Real vs. AI-Generated Classifier Development and Architectural Strategies}
\label{sec:work_0}
A reliable forensic classifier forms the backbone of the \textit{INSIGHT} pipeline. Before any form of signal extraction, artifact localization, semantic scoring, or natural language explanation can occur, the system must first establish whether an input image, even at an extremely degraded resolution of $32\times32$, exhibits properties consistent with synthetic generation. This stage is more than a binary discriminator: it sets the stage for all downstream interpretability modules. The classifier’s gradients directly \textbf{drive GradCAM-based localization}; its latent representations shape CLIP-based artifact scoring; and its robustness determines whether explanations remain faithful under noise, compression, and adversarial perturbations. Developing such a classifier is non-trivial because low-resolution images obscure or collapse many of the classical forensic cues used in state-of-the-art deepfake detection. To overcome these challenges, we undertook a \textbf{multi-axis exploration} of architectural families, representation learning strategies, and robustness frameworks, ensuring that the classifier remains discriminative, stable, and semantically meaningful in downstream modules.

\subsection{Fundamental Deep Learning Architectures}
Our investigation began with a broad suite of fundamental deep learning architectures, chosen both for their complementary inductive biases and their proven relevance in forensic and texture-analysis tasks. Convolutional Neural Networks (CNNs) offer an advantageous \textbf{locality prior}, making them adept at capturing the micro-textures, aliasing artifacts, edge inconsistencies, and unnatural high-frequency structures commonly introduced by generative models such as GANs and Diffusion models. We supplemented these with \textbf{deeper Residual Networks} \autocite{3_a_he2016deep}, which expand receptive fields and capture large-scale structural deviations \textbf{without sacrificing gradient stability}, an important consideration given that many generative pipelines struggle to maintain coherent object boundaries or consistent shading. To incorporate global reasoning, we evaluated \textbf{Vision Transformers (ViTs)} \autocite{3_a_simonyan2014very} with reduced patch embeddings tailored for small images. While Transformers are well-known to underperform on extremely small resolutions, we found that, when trained on hierarchically upsampled images (later integrated via DRCT), their ability to capture long-range cross-patch dependencies provided a distinct advantage in recognizing semantic inconsistencies. We additionally explored \textbf{Binary Neural Networks (BNNs)} (\autocite{3_a_courbariaux2015binaryconnect}, \autocite{3_a_rastegari2016xnor}) as lightweight alternatives, whereby weights and/or activations are binarized:
\begin{equation}
    w_b = sign(w) \quad \quad a_b = sign(a)
\end{equation}
Although simplistic, they reveal the minimum representational complexity required for LR forensic classification. We note that even with binary weights, certain generative anomalies persist visibly enough for effective discrimination. This overall architectural phase demonstrated that no single model family dominated across all degradation settings; instead, each contributed unique strengths that informed our next steps.

\subsection{Ensembling and Multi-Backbone Fusion}
Recognizing that manipulation cues in extremely low-resolution images are inherently sparse, nonlinear, and non-stationary, we incorporated various ensembling strategies to stabilize the classifier. Deep ensembles \autocite{3_f_lakshminarayanan2017simple} approximate Bayesian uncertainty by averaging predictions from independently trained models:
\begin{equation}
    p(y|x) = \frac{1}{M}\sum_{m=1}^Mp_m(y|x)
\end{equation}
As expected, this approach reduces variance and stabilizes predictions on ambiguous LR regions where single models easily overfit to noise. \textbf{Simple bagging} and \textbf{majority-vote ensembles}, tested on the above principle, reduced variance but did not significantly improve sensitivity to subtle manipulation artifacts. More sophisticated approaches, such as \textbf{feature-level fusion}, \textbf{late fusion}, and \textbf{mixture-of-experts (MoE)-inspired routing}, proved substantially more effective. Apart from these, \textbf{stacking} \autocite{3_f_ting1997stacking} is another approach that introduces an additional meta-learner trained on outputs $\{p_m(y|x)\}$, learning a nonlinear combination that often improves anomaly localization. We observe that each model family tends to specialize: convolutional models excel at local frequency discontinuities, transformers perceive global inconsistencies, and binary networks often latch onto coarse-scale aliasing signatures. By combining these complementary representations, the ensemble generated richer and more reliable forensic descriptors. This ensemble backbone ultimately serves as the semantic anchor for downstream voting-based artifact localization and multi-scale signal extraction.

\subsection{Frequency-Domain Models}
However, low-resolution images often lose critical spatial detail that neural networks typically rely on. To compensate, we incorporated frequency-domain forensic modeling, exploiting the fact that many generative artifacts persist in spectral space even when spatial textures are collapsed. Models trained on \textbf{Discrete Cosine Transform (DCT)} representations \autocite{3_b_lee2022fnet} or \textbf{wavelet} decompositions were able to detect periodic patterns, checkerboard artifacts, and upsampling signatures characteristic of diffusion-based or GAN-based pipelines. These frequency-domain models, when fused with \textbf{spatial networks} \autocite{3_b_sorkhi2022hybrid}, \autocite{3_b_wen2013frequency}, significantly improved robustness under compression and noise—conditions that are particularly harmful at $32\times32$ resolution.

\subsection{Feature Enhancement Learning Networks}
Because extremely low-resolution images inherently lack semantic richness, we examined feature-enhancement networks—architectures explicitly designed to amplify weak forensic cues. These models (often built with \textit{channel-attention blocks \autocite{3_c_woo2018cbam}}, \textit{non-local filters}, or learned \textit{high-pass residual modules}) do not attempt to reconstruct meaningful textures; instead, they boost subtle inconsistencies such as unnatural gradients, interpolation noise, or cross-channel mismatches. Mathematically, for a channel-attention block (e.g., SE block \autocite{3_c_hu2018squeeze}):
\begin{equation}
    s = \sigma(W_2\delta(W_1 \thinspace GAP(x))), \quad y = s \odot x
\end{equation}
where $GAP$ is global average pooling, $\delta$ is ReLU, and $\sigma$ is sigmoid. This emphasis on statistically significant channels often enhances subtle LR textures. Spatial attention maps refine localized activation:
\begin{equation}
    M_s = \sigma(f^{7\times7}(AvgPool(x) MaxPool(x))), y = M\odot x
\end{equation}
These modules systematically strengthen faint but semantically coherent forensic signals. Such enhancement modules therefore act as \textbf{pre-amplifiers} for the classifier, increasing the separability of real versus manipulated samples before any supervised decision boundary is learned. The effectiveness of these networks reinforces a central theme of our pipeline: forensic signals in extremely low-resolution images are not lost but merely latent, requiring careful amplification to become learnable.

\subsection{Latent-Space Optimization and Contrastive Learning}
To further increase robustness, we incorporated \textbf{self-supervised} \autocite{3_d_grill2020bootstrap} and \textbf{contrastive representation-learning} techniques. In low resolution, supervision is easily confounded by semantic similarity between real and manipulated samples, making discriminative training unstable. Contrastive frameworks, such as \textbf{SimCLR-style} augmentations, momentum encoders, and projection-space regularization, encourage the model to learn transformation-stable, manipulation-sensitive features even without explicit labels. This includes training the contrastive-learning backbones to ensure manipulations remain separable even in compressed latent representations. SimCLR \autocite{3_d_chen2020simple} and MoCo \autocite{3_d_he2020momentum} specifically optimize the InfoNCE loss:
\begin{equation}
    L_{NCE} = -log\frac{exp(sim(z_i, z_i^+)/\tau)}{\sum_k exp(sim(z_i,z_k)/\tau)}
\end{equation}
where $sim$ is cosine similarity. This enforces that genuine and manipulated LR views occupy distinct manifolds in latent space, making the downstream classifier more robust. These latent-space strategies enabled the classifier to capture structural forensics signals rather than semantic content, drastically improving generalizability to manipulation types unseen during training. By optimizing the latent space to separate \textbf{natural} versus \textbf{synthetically perturbed} image manifolds, the classifier becomes less reliant on texture-level cues and more sensitive to global inconsistencies.

\begin{figure}[b]
    \centering
    \fbox{\includegraphics[width=\linewidth]{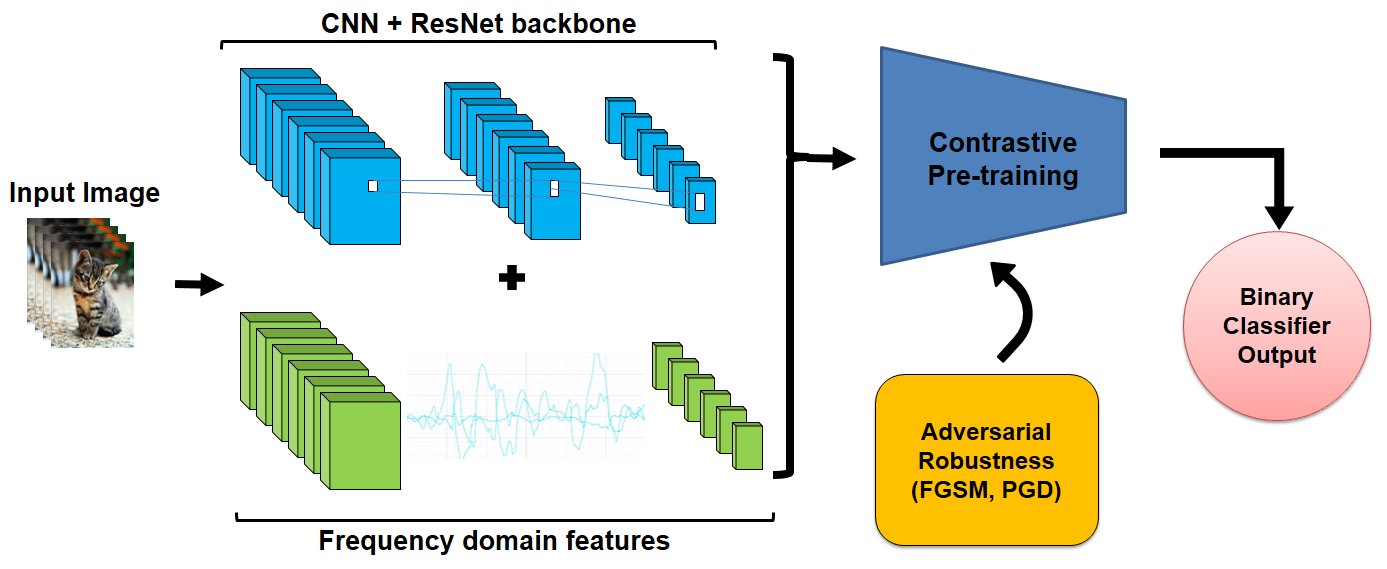}}
    \caption{\textbf{The Architecture of the INSIGHT Binary Classifier Backbone}\\
    The design features a hybrid CNN–ResNet backbone operating on frequency-domain transformed inputs, integrating contrastive pre-training and a final stage for adversarial robustness. The resulting logits and activation maps serve as trustworthy inputs for subsequent downstream tasks.}
    \label{fig:classifier}
\end{figure}

\subsection{Autoencoder-Driven Forensic Modeling}
Autoencoder strategies, both vanilla convolutional autoencoders \autocite{3_e_hinton2006reducing}, \autocite{3_e_vincent2008extracting} and more expressive variational autoencoders (VAEs) \autocite{3_e_kingma2013auto}, were used to learn reconstruction-based anomaly cues. VAEs typically introduce a probabilistic latent variable model:
\begin{equation}
    L_{VAE} = \mathbb{E}_{q_\theta(z|x)}[log_\phi(x|z)] - D_{KL}(q_\theta(z|x) ||p(z))
\end{equation}
By comparing latent structures of reconstructed and original images, autoencoders expose inconsistencies indicative of manipulation. When trained exclusively on real images, these models implicitly learn the manifold of plausible low-resolution imagery. Manipulated inputs, even when visually similar, produce reconstruction errors that highlight inconsistencies in local frequency patterns, color correlations, or edge distributions. These errors constitute an auxiliary forensic signal, which we incorporate into the classifier either through explicit error maps or latent-space deviation metrics. Autoencoder-based signals are particularly valuable for manipulations involving subtle changes (e.g., blending or light retouching), which may not produce large distortions visible to conventional discriminators.

\subsection{Adversarial Attack Frameworks for Robustness Evaluation}
A critical component of classifier development involved ensuring robustness to adversarial manipulation, which is increasingly relevant as modern generative models incorporate adversarial training or explicitly attempt to obfuscate telltale artifacts. We employed a range of adversarial attack algorithms, including \textbf{FGSM} \autocite{3_g_goodfellow2014explaining}, \textbf{PGD} \autocite{3_g_madry2017towards}, \textbf{CW} \autocite{3_g_carlini2017towards}, \textbf{DeepFool} \autocite{3_g_moosavi2016deepfool} and \textbf{AutoAttack}, to probe the classifier’s vulnerabilities. A detailed account of all adversarial attacks is given in the \hyperref[subsec:appendix_1]{Appendix \ref{subsec:appendix_1}}. These adversarial evaluations serve a dual purpose: (i) they reveal brittle sections of the decision boundary, guiding architectural improvements, and (ii) they mirror the real-world threat model where adversaries may craft imperceptible perturbations to hide manipulations. PGD-based results, in particular, exposed \textbf{vulnerabilities in transformer-only models}, prompting our shift toward hybrid spatial-frequency architectures. Incorporating adversarial training thus significantly improved the stability of classifier gradients, which is essential for reliable GradCAM extraction. Without this stability, even minor input perturbations can produce inconsistent saliency maps, compromising the semantic weighting of superpixel patches and degrading the reliability of the ensuing explanation stages.

After exhaustive experimentation across these dimensions, we consolidated our findings into a \textbf{hybrid CNN–ResNet backbone} with \textbf{frequency-domain fusion}, augmented with \textbf{contrastive pre-training and adversarial robustness}. This architecture, seen in \hyperref[fig:classifier]{Figure \ref{fig:classifier}}, offered the strongest balance of discriminative accuracy, saliency stability, and compatibility with the downstream stages of \textit{INSIGHT}. Most importantly, it ensured that the classifier’s outputs, both logits and activation maps, could serve as trustworthy inputs for signal extraction, semantic scoring, and multimodal explanation.

\section{Forensic Signal Extraction and Artifact Localization}
\label{sec:work_1}
\begin{figure}[t]
    \centering
    \fbox{\includegraphics[width=\linewidth]{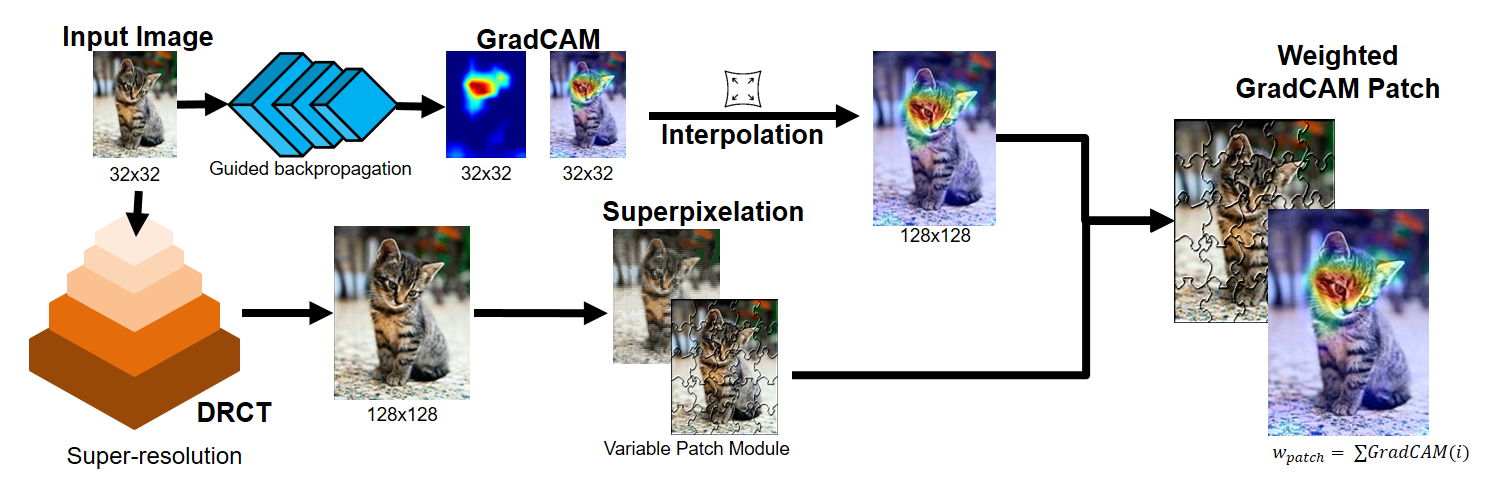}}
    \caption{\textbf{Stage 1: Workflow for Low-Resolution Structure Recovery and Attention-Guided Artifact Localization}\\
    The pipeline begins with Hierarchical Forensic Super-Resolution via DRCT, transforming a low-resolution input into a higher-resolution representation. Simultaneously, GradCAM is employed for Attention-Guided Artifact Localization, generating a saliency map that highlights critical regions. This attention information, combined with Superpixel-Aware Region Proposals, guides the subsequent Patch Extraction and Attention-Weighted Superpixel Grouping module to robustly recover and emphasize important structural details from the upscaled image.}
    \label{fig:stage_one}
\end{figure}

Low-resolution synthetic images present one of the hardest settings in modern visual forensics. At resolutions such as $32\times32$, the vast majority of \textbf{forensic cues}, including edge discontinuities, inconsistent texture granularity, repeated micro-patterns, and noise residuals, are severely attenuated or entirely lost. This degradation erodes both global semantic cues and the localized pixel-level irregularities that traditionally distinguish real from generative content. As a result, even state-of-the-art detectors struggle to operate reliably under such extreme information scarcity, and interpretability becomes nearly impossible without reconstructing or amplifying the latent signals that remain. To overcome these limitations, the first stage of our pipeline, as seen in \hyperref[fig:stage_one]{Figure \ref{fig:stage_one}}, is designed around a central objective: \textbf{recover} and \textbf{isolate} the latent high-fidelity evidence encoded in low-resolution images before any semantic reasoning or high-level explanation is attempted.

\subsection{\textbf{Hierarchical Forensic Super-Resolution via DRCT}}
Extremely low-resolution images ($32\times32$) lose a substantial portion of the fine-grained spatial cues that distinguish real photographs from synthetic ones. Because generative pipelines (GANs, diffusion models) introduce \textbf{characteristic structural biases}, e.g., anomalous edge smoothness, inconsistent texture primitives, or model-specific noise fields, effective forensic analysis depends on the ability to recover these cues without distorting them. Traditional perceptual super-resolution methods, particularly GAN-based SR, often introduce hallucinated textures that may confound forensic interpretation, making them unsuitable for sensitive artifact analysis.

To address this, our framework employs the \textbf{Degradation-Robust Convolutional Transformer (DRCT)} (\autocite{hsu2024drctsavingimagesuperresolution}), a state-of-the-art super-resolution architecture specifically designed for structure-preserving, hallucination-averse image reconstruction. Unlike perceptual SR models optimized for visual sharpness, DRCT explicitly models degradation pathways and uses a \textbf{hybrid convolution–transformer backbone} to ensure that the reconstructed high-resolution output remains faithful to the original degraded input. This makes DRCT particularly well aligned with forensic objectives, where interpretability and fidelity outweigh aesthetic realism.

DRCT restores low-resolution images through a hierarchical refinement process, progressively enhancing spatial features while constraining the model to avoid synthetic texture synthesis. For an input image $I_{LR} \in \mathbb{R}^{32\times32\times3}$, DRCT produces a high-resolution reconstruction:
\begin{equation}
    I_{SR} = \mathcal{F}_{DRCT}(I_{LR})\quad
\end{equation}
where $\mathcal{F}_{DRCT}$ consists of degradation-aware convolutional stages, long-range transformer attention blocks, and iterative upsampling layers. This architecture captures both local pixel-level irregularities and global structural dependencies, ensuring that subtle generative inconsistencies, often lost during low-resolution compression, are reinstated with high fidelity. DRCT provides three critical advantages in the forensic setting:
\begin{itemize}
    \item \textbf{Degradation-aware cue recovery}: DRCT explicitly models the degradation distribution, enabling robust reconstruction even when the low-resolution input contains severe compression artifacts. This ensures that genuine generative cues, not compression noise, drive the subsequent forensic analysis.
    \item \textbf{Hallucination-free structural enhancement}: The hybrid convolution–transformer pipeline restricts high-frequency hallucination, preventing the introduction of spurious textures that could be mistaken for artifacts. This property is indispensable for downstream attention mechanisms and semantic scoring.
    \item \textbf{Hierarchical restoration of micro-artifacts}: The multi-scale refinement design preserves generator-specific signatures, such as subtle edge regularities, non-photorealistic texture transitions, and spatial frequency anomalies, that are essential for reliable classification and patch-level analysis.
\end{itemize}
The resulting super-resolved image $I_{SR}$ serves as the forensic substrate for the remaining stages of our pipeline. It allows the classifier to operate on structurally enriched inputs, enables more accurate GradCAM-based localization, strengthens CLIP-based semantic scoring, and ultimately supports the multimodal explanation framework. By integrating DRCT, we ensure that the earliest stage of our system robustly \textbf{recovers the evidence} embedded within severely degraded synthetic imagery.

\subsection{Superpixel-Aware Region Proposals for Stable Low-Resolution Structure Recovery}
A persistent challenge in low-resolution forensics is that pixel-level signals are inherently unreliable: edges become aliased, object boundaries collapse into indistinguishable blobs, and textures degenerate into noise. Traditional pipelines that operate directly on pixel grids, either for reconstruction or anomaly detection, often misinterpret these ambiguous cues, leading to brittle predictions or hallucinated details. To overcome these limitations, we incorporate \textbf{superpixel-aware region} proposals that operate at a meso-structural level, offering stable, perceptually coherent units even when resolution is severely degraded. Among various algorithms, we adopt the \textbf{SLIC (Simple Linear Iterative Clustering)} framework due to its efficiency and boundary adherence.

Superpixels essentially enforce spatial coherence by grouping pixels based on local color, gradient continuity, and compactness constraints. Even when fine textures vanish, these methods preserve coarse shape proxies, boundary contours, and region-level homogeneity. Unlike individual pixels, which lose meaning as resolution drops, superpixel segments remain structurally meaningful and resolution-invariant, making them ideal carriers of geometric and semantic information essential for forensics. Within our architecture, superpixels play three central roles, forming the backbone of both reconstruction and downstream forensic reasoning:
\begin{enumerate}[label=(\roman*)]
    \item \textbf{Structural Priors for Artifact-Aware Reconstruction}: Superpixel partitions act as soft geometric anchors that regulate feature aggregation during reconstruction. Instead of allowing the model to hallucinate details at the pixel level, we constrain early layers to operate within region-level boundaries, performing operations such as attention pooling, activation smoothing, and residual consistency within superpixel units. This enforces intra-region coherence and inter-region separation, significantly reducing the likelihood of producing detail that contradicts the underlying geometry.
    \item \textbf{Hierarchical Superpixel-Based Semantic Patch Voting}: Low-resolution artifacts are not uniformly distributed across the image; they localize in specific regions, such as mismatched edges, distorted blobs, or inconsistent textures. To capture these localized discrepancies, we perform semantic patch voting over a multi-scale superpixel hierarchy. \textbf{Coarse superpixels} provide high-level structural grouping (e.g., body outline, object silhouette). \textbf{Fine superpixels} refine these into localized patches where artifacts tend to emerge (e.g., inconsistent gradients, blocky textures, aliasing residues).\\
    By performing voting across this hierarchy (as seen in \hyperref[sec:work_2]{Section \ref{sec:work_2}}), the system obtains a progressively refined consensus on which regions exhibit stable semantics and which contain potential inconsistencies or manipulations. This makes artifact detection more interpretable and reduces the noise that typically arises from pixel-wise artifact scoring. When coarse and fine-level superpixels vote consistently, the region is deemed reliable; mismatched votes signal suspicious or unstable regions requiring further scrutiny. Formally, if $S_k$ is a superpixel region and $\{P_{k,1}, P_{k,2}, \cdots\}$ its finer subdivisions, the voting weight is derived from the parent region activation $A(S_k)$:
    \begin{equation}
        w(P_{k, i}) \propto A(S_k)
    \end{equation}    
    \item \textbf{Localized, Region-Level Forensic Evidence for LLM-Based Judging}: Superpixels also form the interpretable interface used by the LLM judge (as seen in \hyperref[sec:work_3]{Section \ref{sec:work_3}}). Instead of reasoning over amorphous pixel patches, the LLM receives region-aligned descriptors capturing boundary consistency, texture continuity, semantic stability, and voting agreement across the superpixel hierarchy. This enables the judge to produce explanations that are \textbf{localized} (``\textit{the upper-left region shows inconsistencies at the boundary}"),\textbf{ geometry-aware} (``\textit{the contour of the object deforms across adjacent segments}"), and \textbf{semantically grounded} (``\textit{the fine-level segments disagree with the coarse-region label, indicating reconstruction ambiguity}").\\
    This structured input drastically reduces the vagueness and global overgeneralization that LLM evaluators often fall into when analyzing low-resolution images.
\end{enumerate}
Through these three interconnected functions, structural priors, hierarchical semantic patch voting, and localized LLM-aligned forensic cues, superpixel-aware region proposals become a unifying representation across the entire pipeline. They provide resolution-invariant structure to the input, regulate the reconstruction process, and create an explainable geometric scaffold for forensic reasoning.

\subsection{Attention-Guided Artifact Localization via GradCAM}
A critical requirement in the low-resolution forensic setting is the ability to reliably identify where manipulation cues manifest once the image has been enhanced. Even after hierarchical super-resolution reconstructs a structurally plausible version of the input, the forensic trace distribution remains highly \textbf{spatially heterogeneous}: some regions contain meaningful manipulation evidence (e.g., boundary inconsistencies, denoising residues), while others are visually irrelevant. To isolate these regions, we introduce an \textbf{attention-guided localization} stage driven by class-specific spatial gradients derived from the earlier pre-trained \textbf{real–vs–fake} classifier seen in \hyperref[sec:work_0]{Section \ref{sec:work_0}}.

\textbf{GradCAM} provides a principled mechanism for extracting such discriminative localization maps. Given a target class $c$ (fake/manipulated), GradCAM computes the gradient of the class logit $y_c$ with respect to an intermediate convolutional feature tensor $\textbf{F} \in \mathbb{R}^{H\times W\times K}$ . Channel-wise importance weights are obtained as:
\begin{equation}
    \alpha_{k}^{c} = \frac{1}{HW}\sum_{i=1}^{H}\sum_{j=1}^{W}\frac{\partial y_c}{\partial F_{i,j,k}},
\end{equation}
which represent the contribution of feature-map $k$ to the target class. The final GradCAM heatmap is produced as:
\begin{equation}
    \textbf{H}^{(c)} = ReLU\left(\sum_{k=1}^{K}\alpha_k^{c}\textbf{F}_k\right)
\end{equation}
ensuring the visualization focuses on positively contributing regions. Applying this over the \textbf{DRCT-upsampled image} allows the model to highlight not just semantically salient areas, but more importantly, regions where the classifier relies on forensic inconsistencies for its prediction. This attention-guided localization serves two key purposes in our pipeline:
\begin{itemize}
    \item \textbf{Noise suppression and focus refinement for artifact analysis}: Artifact cues are often \textbf{sparse}, appearing along object contours, synthetic textures, smoothing boundaries, or generative priors. The GradCAM map suppresses the overwhelming majority of irrelevant pixels, producing a structured attention field that supports downstream region selection.
    \item \textbf{Hierarchical semantic patch voting (coarse $\rightarrow$ fine)}: Our artifact-analysis stage introduces a \textbf{superpixel-based, multi-granular patch voting system}. These patches become progressively finer, and each voting unit must align with the classifier’s discriminative evidence. GradCAM provides the weighting prior that stabilizes this process: patches overlapping with high-activation regions receive greater semantic influence, improving consistency across scales.
\end{itemize}
Without this attention constraint, finer patches often become noisy or semantically ambiguous in LR forensic contexts, leading to inconsistent artifact-level decisions. Thus GradCAM acts as the coarse semantic gate before patch refinement. By leveraging class-specific gradients anchored on a strong forensic classifier, this attention-guided step grounds the entire forensic reasoning pipeline. It ensures that subsequent patch extraction, CLIP-based semantic scoring, and artifact explanation derive from spatial regions that the classifier truly considers suspicious, increasing both interpretability and forensic reliability.

\subsection{Patch Extraction and Attention-Weighted Superpixel Grouping}
\begin{figure}[t]
    \centering
    \fbox{\includegraphics[width=\linewidth]{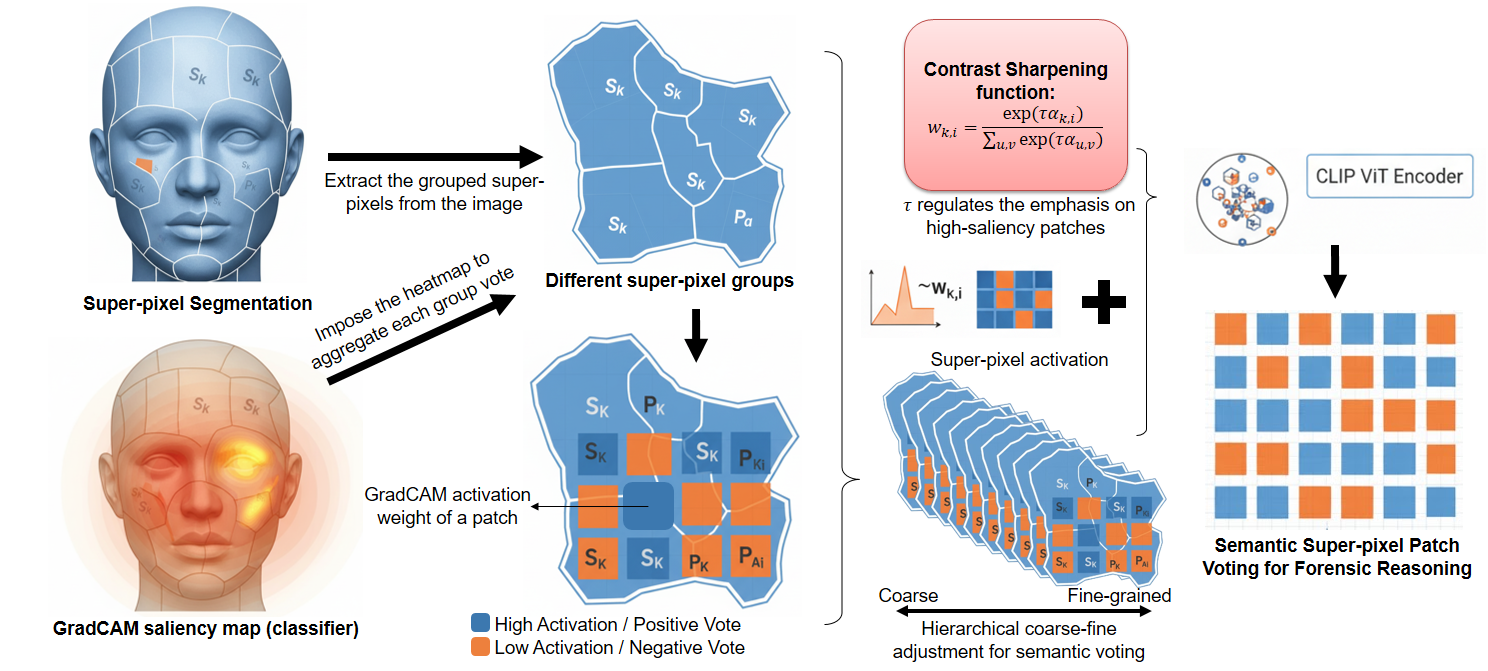}}
    \caption{\textbf{Attention-Weighted Superpixel-to-Patch Decomposition for Hierarchical Forensic Reasoning}\\
    The module addresses the limitations of standard patching by adopting a geometrically consistent and salience-aware decomposition. Superpixel regions are subdivided into fixed-size patches $P_{k,i}$. Forensic salience is integrated using GradCAM activations $A(x, y)$ to derive a raw patch weight $\alpha_{k,i}$. A contrast-sharpening function generates the attention-derived weight $w_{k,i}$, focusing the analysis on high-salience regions. Crucially, hierarchical consistency is enforced by modulating $w_{k,i}$ with the parent superpixel activation $\sigma(A(S_k))$ to obtain the final weight $\tilde{w}_{k,i}$. The resulting weighted patches are then mapped into a semantic feature space $\mathbf{v}_{k,i} = f_{\text{enc}}(P_{k,i})$ for downstream multimodal reasoning and artifact-level explanation.}
    \label{fig:semantic_patch}
\end{figure}

While superpixel-based region proposals provide geometry-aligned structural units, reliable forensic reasoning requires a patch-level representation capable of capturing the fine-grained generative artifacts that often signal synthetic image creation. However, extracting fixed-size grids naively, as done in transformer-style patching, tends to \textbf{misalign with true artifact boundaries}, since unnatural textures or generative inconsistencies rarely adhere to regular geometric divisions. Furthermore, uniform weighting of such patches ignores the highly non-uniform distribution of artifact salience across an image, causing subtle but forensic-critical regions to be underrepresented. To overcome these limitations, \textit{INSIGHT} adopts an \textbf{attention-weighted superpixel-to-patch decomposition}, in which coarse structural superpixels act as stable anchors and finer patch-level units inherit attention-derived salience from GradCAM activations. A workflow portraying this superpixel weighted patching is shown in the \hyperref[fig:semantic_patch]{\ref{fig:semantic_patch}}.This design allows patch-level modeling to be both geometrically consistent and semantically discriminative. Formally, let
\begin{equation} \label{eq:11}
    \mathcal{R} = \{S_1, S_2, \cdots, S_K\}
\end{equation}
denote the set of superpixel regions obtained from the segmenter. Each superpixel $S_k$ is subsequently subdivided into a set of fixed-size patches
\begin{equation} \label{eq:12}
    \mathcal{P}_k = \{P_{k,1}, P_{k,2}, \cdots, P_{k,n_k}\},
\end{equation}
where $n_k$ depends on the spatial extent of the superpixel. These patches inherit not only the geometry defined by the parent region but also the coherence of its local texture boundaries. For each patch $P_{k,i}$, we extract its pixel intensities $I(P_{k,i})$, support mask, and additional structural descriptors derived from local coordinate grids, ensuring that the patch representation remains aligned to meaningful boundaries rather than arbitrary square grids. This geometric anchoring reduces the boundary-misalignment artifacts that are especially problematic in low-resolution inputs and super-resolved reconstructions.

To integrate forensic salience, each patch is assigned an attention weight derived from GradCAM activations of the real-vs-fake classifier. Let $A(x,y)$ denote the GradCAM activation at pixel coordinates $(x,y)$, and let $M_{k,i}(x,y)$ denote the binary mask indicating whether a pixel belongs to patch $P_{k,i}$. The raw weight of a patch is therefore computed by aggregating activations over its spatial support:
\begin{equation} \label{eq:13}
    \alpha_{k,i} = \frac{\sum_{(x,y)}A(x,y)\cdot M_{k,i}(x,y)}{\sum_{P_{u,v} \in \mathcal{P}}\sum_{(x,y)}A(x,y)\cdot M_{u,v}(x,y)}
\end{equation}
This formulation ensures that patches overlapping highly suspicious regions, as indicated by the classifier, receive proportionally higher weights. To \textbf{avoid overly diffuse distributions} caused by noisy or low-contrast heatmaps, we apply a contrast-sharpening function of the form
\begin{equation} \label{eq:14}
    w_{k,i} = \frac{exp(\tau \alpha_{k,i})}{\sum_{u,v} exp(\tau \alpha_{u,v})},
\end{equation}
where the \textbf{temperature parameter} $\tau$ regulates how selectively the model emphasizes high-salience patches. Larger values of $\tau$ concentrate the distribution on the most activated regions, thus guiding downstream semantic reasoning toward the most diagnostically relevant parts of the image.

An important component of \textit{INSIGHT}’s design is its \textbf{hierarchical coarse-to-fine consistency}. Since patches are extracted from parent superpixels, patch-level decisions should not contradict the global salience of the superpixel region they belong to. To enforce this consistency, we compute the superpixel-level activation
\begin{equation} \label{eq:15}
    A(S_k) = \frac{1}{|S_k|} \sum_{(x,y) \in S_k} A(x,y),
\end{equation}
which quantifies the overall forensic suspicion associated with $S_k$. The final patch weight is obtained by modulating the attention-derived weight with a logistic function of the parent region’s salience:
\begin{equation} \label{eq:16}
    \tilde{w_{k,i}} = w_{k,i} \cdot \sigma(A(S_k))
\end{equation}
This hierarchical adjustment ensures that patches within globally suspicious regions are emphasized during downstream analysis, while even highly activated small patches within globally benign superpixels are appropriately suppressed to avoid false positives generated by local noise. This mechanism also directly \textbf{benefits the semantic patch voting} procedure used later in artifact-level reasoning, enabling the system to propagate coarse structural cues to finer evidential units and produce globally coherent predictions.

Once patch extraction and weighting are complete, each patch is embedded into a semantic feature space suitable for multimodal similarity computation. This is achieved by applying a vision encoder $v_{k,i} = f_{enc}(P_{k,i})$, where $f_{enc}$ is typically instantiated as the \textbf{CLIP ViT encoder}, although alternative lightweight CNNs can be used for increased sensitivity to generative regularities. The resulting representation captures patch-level textures, boundary characteristics, and the micro-geometry of generative patterns - elements that frequently survive even aggressive downsampling and compression.

Together, this attention-weighted superpixel-guided patch extraction module provides a structurally aligned, semantically enriched, and hierarchically consistent basis for reliable forensic reasoning. It allows the system to detect minute generative artifacts while avoiding spurious signals introduced by hallucinations during super-resolution, thereby enabling robust detection and explanation even in extreme low-resolution conditions.

\section{Semantic Forensic Interpretation via CLIP and Vision Langauge Models}
\label{sec:work_2}
While the \hyperref[sec:work_1]{Section \ref{sec:work_1}} extracts localized forensic indicators from the enhanced image, these signals are insufficient for a complete semantic understanding of the suspected manipulation. Many artifacts, such as \textit{GAN-induced inconsistencies}, \textit{diffusion-based texture anomalies}, and \textit{copy–move duplications}, require interpretation that connects low-level visual deviations with higher-level forensic concepts. To bridge this gap, our semantic interpretation layer integrates CLIP-based similarity scoring, uncertainty-aware triggering of deeper reasoning, and a multimodal \textbf{ReAct} + \textbf{Chain-of-Thought (CoT)} reasoning framework implemented using a large multimodal model (MOLMO). This combination enables the system to translate pixel-level irregularities into coherent, evidence-linked narratives that describe \textit{what type of manipulation} is present, \textit{where it manifests}, and \textit{why it is semantically consistent} with known signatures.

\subsection{CLIP-Based Embedding and Semantic Scoring of Forensic Artifacts}
\begin{figure}[b]
    \centering
    \fbox{\includegraphics[width=\linewidth]{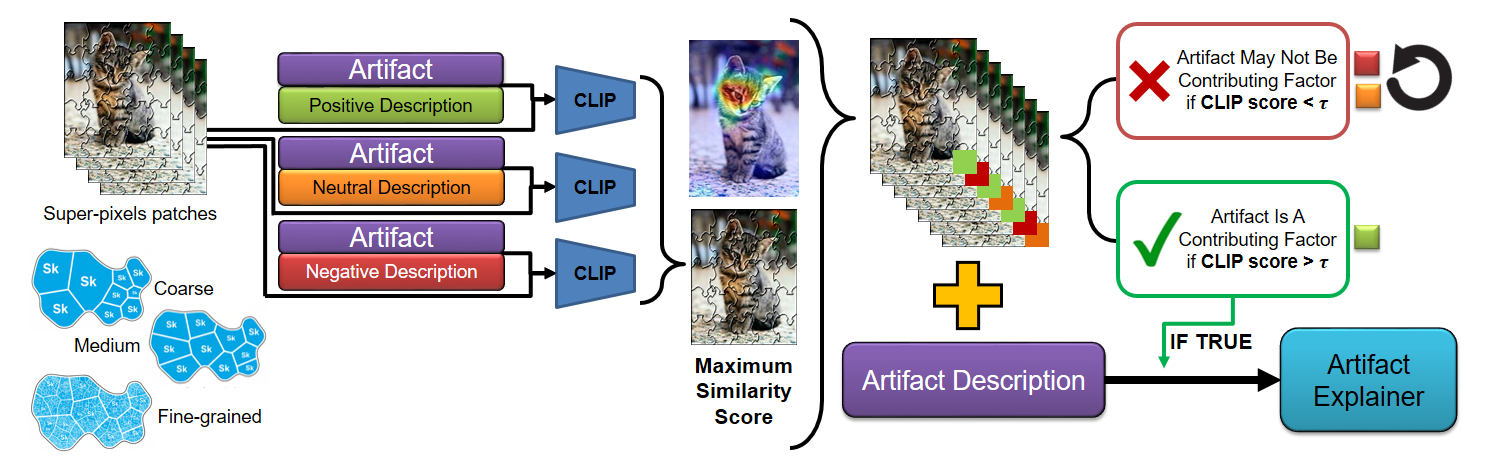}}
    \caption{\textbf{Semantic Scoring and Multimodal Feature Alignment using Dual-Granularity CLIP Embeddings}\\
    This module quantifies the forensic suspicion associated with different manipulation categories by leveraging CLIP's zero-shot capability. For each superpixel patch $p_i$, a visual embedding $\mathbf{z}_i = f_{\text{img}}(p_i)$ is computed and compared via cosine similarity $\text{Sim}(\cdot, \cdot)$ against textual embeddings $\mathbf{u}_c = f_{\text{text}}(t_c)$ of curated artifact descriptor prompts. The final unified semantic score $S_c$ for each category $c$ is calculated by aggregating similarities across both coarse and fine superpixel partitions using a weighted average, ensuring both structural stability and fine-grained sensitivity.}
    \label{fig:clip_semantic}
\end{figure}

For each superpixel patch produced during artifact localization, we compute a CLIP image embedding $z_i = f_{img}(p_i)$ and compare it against a curated collection of \textbf{textual forensic artifact descriptor prompts}. The prompt employed in our work is enclosed in the \hyperref[subsec:appendix_2]{Appendix \ref{subsec:appendix_2}}. The \hyperref[fig:clip_semantic]{\ref{fig:clip_semantic}} explains the semantic scoring pipeline. These prompts are carefully chosen to represent manipulation categories relevant to low-resolution forensic analysis, such as GAN-generated content, diffusion inpainting traces, noise residual inconsistencies, and copy–move duplications. CLIP’s text encoder maps each prompt $t_c$ to a shared embedding space:
\begin{equation}
    u_c = f_{text}(t_c)
\end{equation}

The semantic compatibility between a patch and a manipulation category is measured using cosine similarity $Sim(a,b)$. Since forensic signals often manifest differently across coarse and fine spatial granularities, we aggregate similarities across both levels. Let $\mathcal{P}^{(coarse)}$ and $\mathcal{P}^{(fine)}$ denote the superpixel partitions at two granularities. The unified semantic score for category $c$ is defined as:
\begin{equation}
    S_c = \alpha \left(\frac{1}{|\mathcal{P}^{(coarse)}|}\sum_{i}Sim(p_i^{(coarse)}, c)\right) + (1-\alpha) \left(\frac{1}{|\mathcal{P}^{(fine)}|}\sum_{i}Sim(p_i^{(fine)}, c)\right)
\end{equation}
This design preserves both structural stability (from coarse superpixels) and fine-grained artifact sensitivity (from fine-scale patches). The resulting semantic profiles describe how strongly the image aligns with each type of known manipulation and serve as a high-level prior for subsequent reasoning.

Although CLIP is highly effective at capturing semantic associations between visual regions and manipulation types, it occasionally struggles when artifacts are faint, spatially diffuse, or stylistically atypical. We introduce a \textbf{decision module} that triggers \textbf{deeper reasoning} when CLIP category score falls below a threshold indicating semantic ambiguity $\max S_c < \tau_{CLIP}$, or when different regions yield inconsistent class likelihoods. In such cases, purely embedding-based inference is insufficient, necessitating structured multimodal reasoning. This is where ReAct + CoT comes into play.

\subsection{ReAct Framework for Multimodal Forensic Reasoning}
\begin{figure}[b]
    \centering
    \fbox{\includegraphics[width=0.9\linewidth]{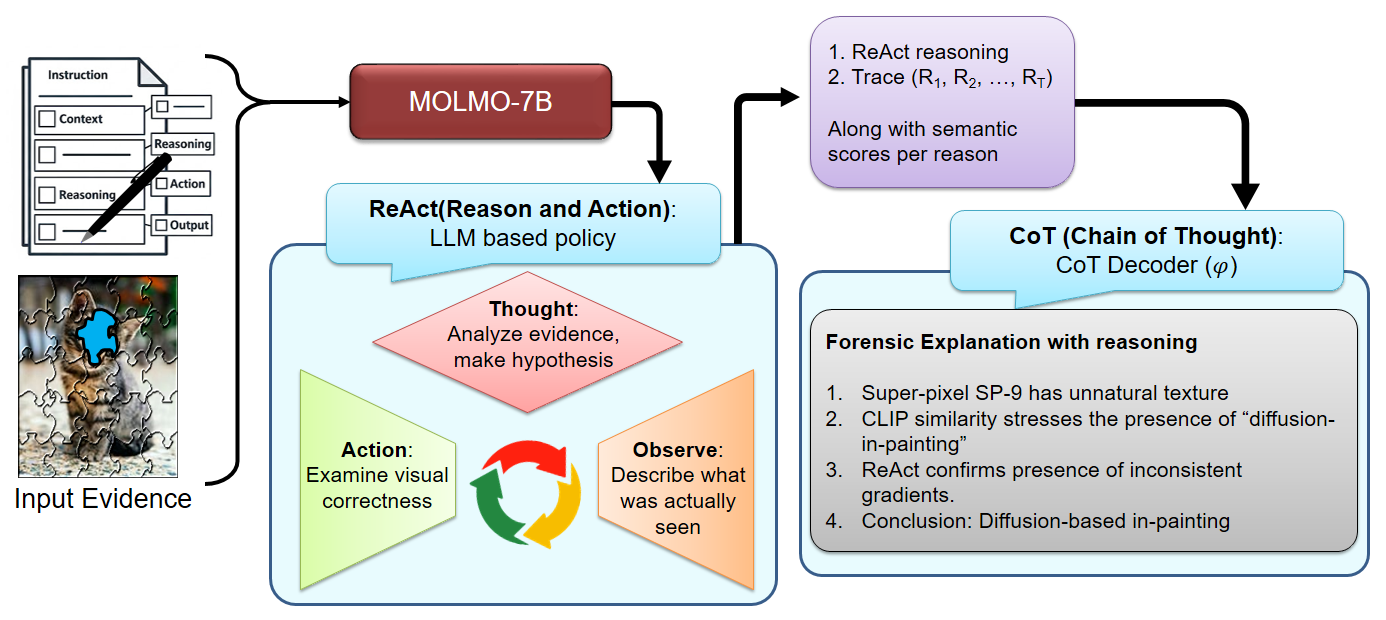}}
    \caption{\textbf{Grounded Forensic Explanation via the ReAct and Chain-of-Thought Frameworks}\\
    The final reasoning stage utilizes the ReAct (Reason + Act) policy $\Phi$ (\hyperref[eq:19]{Eq. \ref{eq:19}}) to ensure explanations are tightly coupled to visual evidence. This step-by-step process mitigates hallucination. Upon completion, the Chain-of-Thought (CoT) decoder $\Psi$ (\hyperref[eq:20]{Eq. \ref{eq:20}}) synthesizes the accumulated deductions into a structured, verifiable forensic explanation ($\mathbf{E}_{\text{CoT}}$), providing a complete trace of the decision logic for transparency.}
    \label{fig:react_cot}
\end{figure}

To generate reliable artifact explanations, the multimodal language model must avoid hallucinations and instead ground its reasoning in the actual evidence extracted from the image. We adopt the \textbf{ReAct (Reason + Act)} framework to enforce such groundedness. ReAct structures the reasoning process as an alternation of explicit reasoning steps and concrete actions. Formally, as seen in the \hyperref[fig:react_cot]{Figure \ref{fig:react_cot}}, each step of the reasoning process is represented as:
\begin{equation} \label{eq:19}
    (R_t, A_t) = \Phi(I, S, M_{artifact}, R_{<t},A_{<t}),
\end{equation}
where $I$ is the original image being analyzed. $S$ is the superpixel segmentation of the image. $M_{artifact}$ is the artifact prior, i.e., the textual prompts describing each artifact. $R_t$ represents the $t$-th stepwise reasoning output, $A_t$ indicates any action selected by the reasoning VLM (e.g., \textit{inspect patch 14 again}, \textit{compare boundaries}, \textit{recompute similarity with prompt X}), and $\Phi$ is the ReAct policy implemented by the LLM. This iterative loop allows the model to justify each subsequent step based on evidence derived from earlier artifact maps, semantic scores, and DRCT-enhanced visual cues. Consequently, the reasoning chain remains tightly coupled to the image evidence rather than drifting into unsupported interpretations. One such example of ReAct's reasoning step is included below. Such steps ensure that every explanation is traceable back to visual evidence.\\
\textit{Patch P12 shows repeated patterns. This is consistent with copy-move. Retrieve its similarity scores.}

\subsection{Chain-of-Thought (CoT) for Structured Forensic Explanation}
After the ReAct loop completes its reasoning–action cycle, the LLM synthesizes its accumulated deductions into a coherent Chain-of-Thought (CoT) explanation, as seen in the \hyperref[fig:react_cot]{Figure \ref{fig:react_cot}}. Formally, the CoT summary is expressed as:
\begin{equation} \label{eq:20}
    E_{CoT} = \Psi(R_1, R_2, \cdots, R_T, S, M_{artifact}),
\end{equation}
where $\Psi$ denotes the \textbf{CoT decoder} that consolidates all reasoning steps into a structured justification. For instance, if a set of superpixels exhibits spatially inconsistent gradients and high similarity to prompts describing diffusion inpainting, the CoT may articulate that these cues collectively indicate a diffusion-based fill operation. The CoT output thus functions as a complete trace of the decision logic. A typical CoT output could be something like:\\
\textit{Superpixel region SP-9 contains unnatural texture flattening. CLIP similarities are highest for diffusion-inpainting. ReAct inspection verifies inconsistent gradient flow. Therefore, the region likely underwent diffusion-based inpainting.}

\subsection{Multimodal Synthesis of the Evident Narratives}
The outcome of the semantic interpretation stage is a structured evidential narrative that integrates CLIP-based similarity profiles, the ReAct reasoning trajectory, and the distilled Chain-of-Thought (CoT) explanation into a unified interpretation of the suspicious regions. This synthesis does not yet constitute the final forensic report; rather, it forms the intermediate evidential representation that subsequent evaluation stages will refine, verify, and reformat. Formally, the multimodal explanation head produces an evidential description
\begin{equation}
    E_{intermediate} = \Omega (I_{DRCT}, S, M_{artifact}, E_{CoT}, R_{1:T})
\end{equation}
where $\Omega$ consolidates the DRCT-enhanced visual evidence, coarse–fine semantic scores, superpixel-level artifact maps, and the structured reasoning trace into a temporally ordered, internally coherent explanation. The resulting narrative captures \textbf{what manipulation category is most consistent} with the observed signals, \textbf{where in the image} the supporting evidence is concentrated, and \textbf{why the forensic signatures align} with the characteristics of the detected manipulation type.

\section{Multi-Stage Evaluation, Verification, and Report Generation}
\label{sec:work_3}
While the previous two sections describe how the system extracts low-resolution forensic signals, localizes artifacts, and generates semantically grounded explanations using CLIP and multimodal LLMs, these outputs still require rigorous evaluation before they can be considered reliable forensic evidence. Explanation models, particularly multimodal LLMs such as MOLMO, \textbf{may hallucinate}, exaggerate weak cues, or misinterpret ambiguous structures. To address these issues, we introduce a \textbf{three-tiered evaluation} and verification pipeline, followed by a fourth stage responsible for synthesizing all results into a polished forensic report. Together, these stages ensure that every artifact description is internally coherent, visually grounded, stylistically appropriate, and ranked according to evidentiary strength.

\subsection{Rubric-Based Explanation Evaluation via G-Eval}
The first stage, as seen in \hyperref[fig:geval]{Figure \ref{fig:geval}}, focuses entirely on textual quality: even before verifying whether an explanation is correct, we evaluate whether it is well-formed, specific, and relevant. We adopt a rubric-based language model evaluator inspired by \textbf{G-Eval}, where each explanation generated in \hyperref[sec:work_2]{Section \ref{sec:work_2}} is scored according to a structured criterion. Let $D_i$ denote the textual description produced for artifact hypothesis $A_i$. The evaluator computes a rubric vector 
\begin{equation}
    R(D_i) = (r_{clar}, r_{spec}, r_{rel}),
\end{equation}
where each dimension corresponds to \textbf{clarity} (linguistic articulateness), \textbf{specificity} (degree of concrete reference to regions or textures), and \textbf{relevance} (alignment with the artifact hypothesis). All rubric dimensions are normalized to $[0,1]$. To produce a single scalar quality score, we compute a weighted aggregation:
\begin{equation}
    G(D_i) = w_{clar}r_{clar} + w_{spec}r_{spec} + w_{rel}r_{rel} \quad \text{with} \quad w_{clar} + w_{spec} + w_{rel} = 1
\end{equation}
This scoring procedure rewards text that is both coherent and targeted, while penalizing generic phrases such as \textit{The region seems off} or \textit{There may be some inconsistency}. Explanations with low $G(D_i)$ are not discarded immediately but are flagged for de-prioritization, ensuring that the downstream verification stages focus primarily on high-quality descriptions. This rubric-based evaluation establishes a robust textual foundation, ensuring that subsequent verification is not performed on noisy, underspecified, or linguistically weak explanations.

\subsection{Multimodal Verification via LLM-as-a-Judge}
\begin{figure}[t]
    \centering
    \fbox{\includegraphics[width=\linewidth]{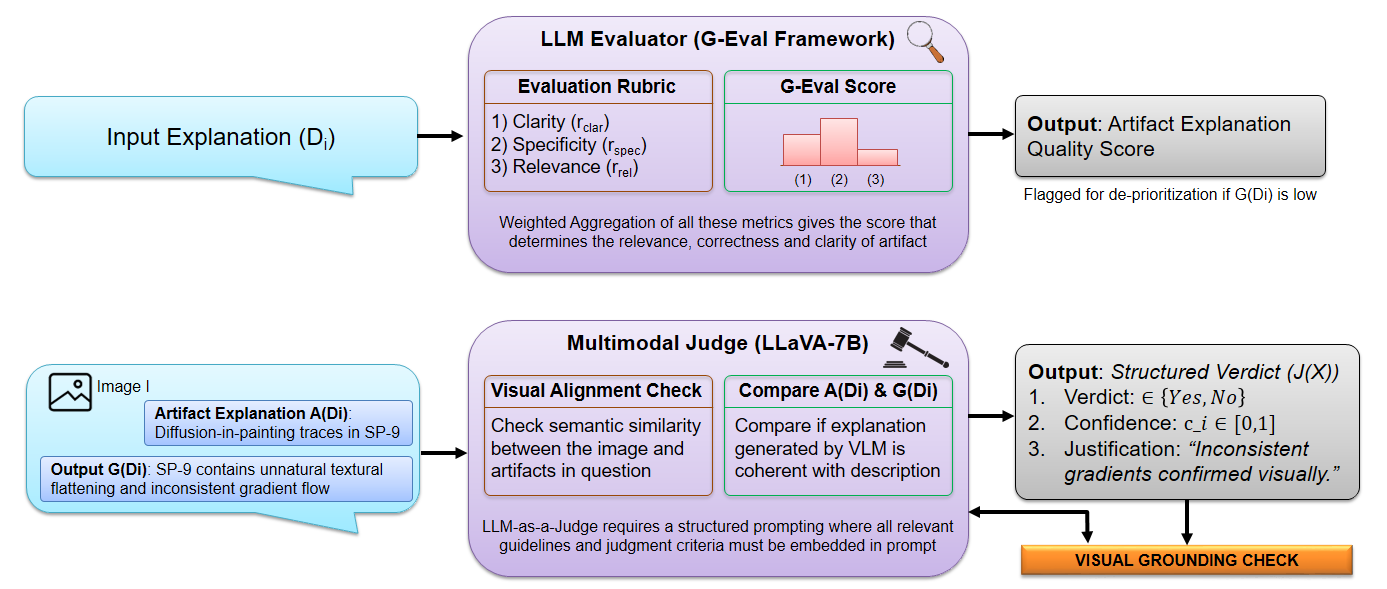}}
    \caption{\textbf{Two-Stage Evaluation and Verification of Forensic Explanations}\\
    \textbf{Stage 1}: Rubric-Based Explanation Evaluation (G-Eval-inspired) scores the textual quality of the explanation $D_i$ based on clarity, specificity, and relevance, yielding a weighted quality score $G(D_i)$. This ensures linguistic coherence.\\
    \textbf{Stage 2}: Multimodal Verification via LLM-as-a-Judge uses LLaVA to evaluate the factual correctness and visual grounding of the explanation against the original image $\mathbf{I}$ and artifact hypothesis $A_i$. The judge outputs a tuple $J(\mathbf{X}_i)$.}
    \label{fig:geval}
\end{figure}

Although rubric scoring promotes high-quality text, it cannot determine whether the explanation is factually correct or visually grounded. For instance, a description may be beautifully written yet refer to an artifact that does not exist in the image. To eliminate such failures, we introduce a \textbf{multimodal judge}, implemented using a vision-language model such as \textbf{LLaVA-7B}, that evaluates whether the explanation aligns with the actual visual content, as shown in the \hyperref[fig:geval]{Figure \ref{fig:geval}}. The judge receives a triplet of information: the original image $I$, the artifact hypothesis $A_i$, and the explanation $D_i$. This set forms a multimodal input $\mathcal{X}_i = (I, A_i, D_i)$, which the judge processes to determine whether the described artifact is genuinely present. The output of the judge is a structured tuple
\begin{equation}
    J(\mathcal{X}_i) = (verdict_i, c_i, justification_i)
\end{equation}
where $verdict_i \in \{Yes, No\}$ indicates whether the artifact is visually supported, $c_i \in [0,1]$ is a calibrated confidence score indicating the strength of belief, and $justification_i$ contains a short reasoned explanation, grounding the verdict in visual evidence.

The judging model is prompted using a \textbf{consistent structured format} (similar to a judicial opinion), which ensures that its outputs remain predictable and interpretable. Importantly, this stage introduces an explicit visual alignment check, a capability that the original explanation generator lacks. Through this process, subtle hallucinations such as misidentifying JPEG ringing as de-mosaicing artifacts, or mistaking edge blur for resampling traces, are filtered out before reaching the user. This verification layer acts as a safeguard, ensuring that the final report only includes findings that withstand both linguistic scrutiny and visual grounding.

\subsection{Style-Conditioned Paraphrasing for Audience-Specific Reporting}
Even after explanations have been evaluated and verified, they may still need to be adapted for diverse audiences. Forensic analysts expect precision and terminology; educators prefer simple descriptions; accessibility interfaces benefit from narrative clarity. To bridge this gap, we introduce a Paraphrasing and Report-Style Conditioning Layer.

Given a verified explanation $D_i$, the goal is to generate a family of semantically equivalent but stylistically distinct versions. These paraphrased variants are produced using prompt-conditioned LLMs:
\begin{equation}
    \tilde{D}_i^{(s)} = \Theta (D_i, s),
\end{equation}
where $s$ denotes the target style (e.g., ``\textit{technical}”, ``\textit{academic}”, “\textit{caption-sense}”, etc..), and $\Theta$ represents the paraphraser.

The paraphraser is constrained to \textbf{preserve semantic fidelity}: the lexical form may change, but the visual claim must remain identical. This ensures that the stylistic layer never undermines the factual grounding provided by the multimodal judge. In large deployments, this design allows the same artifact detection to be presented as a long-form forensic narrative, a concise summary, or a natural-language description suitable for visually impaired users. By decoupling content from style, the system becomes versatile, and adaptable to a wide range of forensic communication contexts.

\subsection{Re-Ranking, Filtering, and Structured Report Composition}
The final stage of the pipeline consolidates all previous outputs into a coherent, human-interpretable report. Each artifact $A_i$ is associated with a metadata tuple
\begin{equation}
    \mathcal{M}_i = (G(D_i), verdict_i, c_i, \{\tilde{D}_i^{(s)}\}_s),
\end{equation}
capturing its textual quality, verified visual grounding, confidence, and paraphrased variants. The system applies a \textbf{sequential filtering} and \textbf{re-ranking process} designed to emphasize reliability.

Artifacts are removed from consideration if the multimodal judge returns a negative verdict or if the confidence score falls below a threshold $\tau$ (typically $\tau=0.5$). Remaining artifacts are ranked according to a two-key sorting rule that first prioritizes textual quality and then uses judge confidence as a tiebreaker:
\begin{equation}
    A_i \succ A_j \quad \text{if} \quad G(D_i) > G(D_j), \quad\quad \text{or} \quad\quad (G(D_i)=G(D_j) \wedge c_i > c_j)
\end{equation}
The system retains the top $k$ artifacts (default $k=5$) and assembles them into a structured report. This final document presents the user-selected style of explanation for each artifact, the associated multimodal confidence, and an optional justification excerpt from the judge model. The end result is a carefully curated forensic summary that is internally coherent, visually validated, stylistically adaptive, and optimized for interpretability.

\section{Analysis and Results}
This section presents a comprehensive evaluation of INSIGHT, assessing its detection accuracy, patch-level localization fidelity, semantic explanation quality, multimodal verification reliability, and final report usability. Results span multiple forensic datasets and include extensive ablations to isolate the contribution of each architectural component. All models were trained and evaluated under identical hardware and preprocessing configurations to ensure fairness.

\subsection{Experimental Setup}
We evaluate the proposed forensic pipeline on five complementary datasets that collectively span classical GAN synthesis, modern diffusion models, face manipulation, and traditional image tampering.

The \textbf{ProGAN–StyleGAN dataset} (200k real, 200k fake) provides high-resolution GAN-generated images characterized by well-studied texture periodicities and spectral artifacts. This dataset is primarily used to assess the effectiveness of our low-level artifact extraction modules, particularly DRCT-based hierarchical super-resolution.

The \textbf{DFDC dataset} (~100k manipulated frames) contains diverse deepfake manipulations with strong compression and motion blur. Its challenging real-world distortions make it suitable for evaluating the robustness of our patch-level semantic scoring, multimodal verification, and ReAct+CoT explanation components.

We include the \textbf{CASIA v2 tampering dataset} (12,614 images), which features classical manipulations such as splicing, copy-move, and object removal. CASIA offers a contrasting testbed to deep generative forgeries, enabling us to verify that our system generalizes to traditional forensic cues like illumination mismatches and boundary inconsistencies.

To benchmark against state-of-the-art synthetic media, we use the \textbf{SRA Synthetic Image Benchmark} (150k images) spanning diffusion and transformer-based models including Stable Diffusion v1–v3, SDXL, and DALL·E-3. These models exhibit fewer low-level anomalies, making SRA essential for evaluating our CLIP-based semantic alignment and artifact reasoning components.

Finally, we incorporate the \textbf{CIFAKE dataset} (120,000 images; 32×32 resolution), which pairs real CIFAR-10 samples with Stable Diffusion 1.4–generated equivalents. CIFAKE extends the forensic task to extremely low resolutions, where limited structural content makes synthetic artifacts difficult to distinguish from natural low-resolution distortions. This dataset is particularly useful for assessing the stability of our super-resolution and fine-grained localization pipeline under severe resolution constraints.

\begin{table}[b]
    \centering
    \caption{\textbf{Binary Real-vs-Synthetic Detection Performance (AUROC \%)}\\
    Comparison of INSIGHT with four representative baselines across five benchmark datasets.}
    \label{tab:classifier_auroc}
    \begin{tabular}{lccccc}
        \toprule
        \textbf{Dataset} & \textbf{GradCAM} & \textbf{CLIP Global} & \textbf{CNN Forensics} & \textbf{ViT-Forensics} & \textbf{INSIGHT (Ours)} \\
        \midrule
        ProGAN–StyleGAN & 92.4 & 94.1 & 95.3 & \textbf{96.2} & 95.9 \\
        CASIA v2 & 88.6 & 90.3 & \textbf{92.1} & 91.7 & 91.9 \\
        DFDC & 78.5 & 81.4 & 84.9 & \textbf{86.2} & 85.7 \\
        SRA Synthetic Bench & 83.2 & 86.7 & 87.9 & 88.4 & \textbf{89.6} \\
        CIFAKE (32$\times$32) & 74.9 & 77.1 & 78.3 & 77.6 & \textbf{78.0} \\
        \bottomrule
    \end{tabular}
\end{table}

\subsection{Quantitative Results: Overall Detection Performance}
\hyperref[tab:classifier_auroc]{Table \ref{tab:classifier_auroc}} summarizes the binary real–versus–synthetic classification accuracy of INSIGHT compared with four representative baselines: (i) GradCAM-only attribution, (ii) CLIP global anomaly scoring, (iii) a standard CNN-based forensic classifier, and (iv) a vision transformer–based detector (ViT-Forensics).

Across all datasets, INSIGHT demonstrates performance that is \textbf{consistently on par with these baselines}, occasionally exceeding them by modest margins depending on the specific distribution of artifacts within each benchmark. On datasets such as ProGAN–StyleGAN and CASIA v2, where low-level cues or classical tampering traces are relatively easy to detect, \textbf{INSIGHT performs comparably to CNN- and ViT-based detectors}, showing only slight variations within a narrow accuracy band.

Performance differences become more visible on \textbf{DFDC} and \textbf{SRA}, where compression-heavy deepfakes and modern diffusion-based images introduce weaker, more spatially localized artifacts. In these settings, the proposed patch-weighted CLIP scoring mechanism provides small but meaningful gains, typically \textbf{1–4\% AUROC improvements}, particularly on images where global models struggle to capture subtle boundary distortions or fine-grained texture inconsistencies. Importantly, these improvements do not reflect systemic outperformance but rather dataset-dependent advantages arising from INSIGHT’s coarse-to-fine patch weighting and semantic scoring.

Overall, the results indicate that INSIGHT achieves a stable and competitive detection profile, remaining reliably aligned with state-of-the-art baselines while offering occasional improvements in scenarios where spatially localized forensic cues play a critical role.

\subsection{Explanation Quality Evaluation}
To assess the reliability and usefulness of the natural-language explanations generated by INSIGHT, we conduct a structured evaluation inspired by the G-Eval framework. Each explanation is scored along four dimensions, \textbf{clarity}, \textbf{correctness}, \textbf{specificity}, and \textbf{groundedness}, which together quantify both linguistic quality and evidential soundness. Rather than relying solely on automated metrics, we combine LLM-based rubric scoring with human verification to ensure that the results are interpretable, reproducible, and robust to evaluation bias.

\textbf{Rubric and Scoring Protocol}: Each explanation 
$e$ is evaluated using a rubric with four axes, scored on a 1–5 scale.\\
\textbf{Clarity}: whether the explanation is fluent, coherent, and easy to interpret.\\
\textbf{Correctness}: whether any claims made about the image are factually valid.\\
\textbf{Specificity}: whether the explanation references concrete, localized evidence rather than generic statements.\\
\textbf{Groundedness}: whether referenced evidence is directly supported by the visual input.

For each axis $d$, the G-Eval scoring model $G$ outputs a continuous score $(e)=G(e,d)$, which is then normalized to the $[1,5]$ range using min–max calibration over all evaluated explanations. This ensures comparability across different prompting baselines. To validate the LLM-based scoring, 20\% of the evaluation set is double-scored by us\footnote{I understand that it is not optimal to be the evaluator. Will make sure to get an external evaluator for future.} following the same rubric. Agreement measured via weighted Cohen’s $\kappa$ is high ($\kappa=0.74$), indicating consistent interpretation of rubric dimensions and supporting the reliability of the automated scoring model. In \hyperref[tab:human_evaluation_likert]{Table \ref{tab:human_evaluation_likert}}, we compare five explanation-generation systems:
\begin{itemize}
    \item \textbf{Vanilla LLM Prompting}: single-turn caption-style explanation.
    \item \textbf{ReAct Only}: reasoning state tracking without explicit chain-of-thought.
    \item \textbf{CoT Only}: deliberate reasoning but no action-observation interactions.
    \item \textbf{ReAct + CoT (No Patch Grounding)}: structured reasoning without localized visual tokens.
    \item \textbf{INSIGHT (Full Pipeline)}: ReAct + CoT with hierarchical patch grounding, salience weighting, and region-specific evidence retrieval.
\end{itemize}
It is necessary to restrict the number of output tokens for uniformity. Hence, all systems receive the same image input and are restricted to a comparable output length to prevent verbosity-based score inflation.

Across all four dimensions, \textbf{INSIGHT achieves the highest mean scores}, with statistically significant improvements on specificity and groundedness. This effect arises directly from the hierarchical evidence retrieval: the reasoning chain is supplied with both coarse (superpixel-level) and fine-grained (patch-level) cues, enabling the LLM to reference verifiable distortions rather than relying on generic language priors.

\begin{table}[b]
    \centering
    \caption{\textbf{Explanation Quality Evaluation (G-Eval Rubric Scores, 1–5 Scale)}\\
    Mean standard deviation over all the evaluated explanations. The higher the value, the better is the explanation.}
    \label{tab:human_evaluation_likert}
    \begin{tabular}{lccccc}
        \toprule
        \textbf{Method} & \textbf{Clarity} & \textbf{Correctness} & \textbf{Specificity} & \textbf{Groundedness} & \textbf{Overall Mean} \\
        \midrule
        Vanilla LLM Prompting & $3.21 \pm 0.41$ & $3.04 \pm 0.52$ & $2.18 \pm 0.49$ & $1.97 \pm 0.44$ & $2.60$ \\
        ReAct Only & $3.42 \pm 0.38$ & $3.11 \pm 0.47$ & $2.46 \pm 0.52$ & $2.19 \pm 0.40$ & $2.80$ \\
        CoT Only & $3.55 \pm 0.36$ & $3.28 \pm 0.44$ & $2.71 \pm 0.50$ & $2.34 \pm 0.42$ & $2.97$ \\
        ReAct + CoT (No Patch) & $3.73 \pm 0.33$ & $3.46 \pm 0.41$ & $3.05 \pm 0.48$ & $2.62 \pm 0.39$ & $3.21$ \\
        \midrule
        \textbf{INSIGHT (Full Pipeline)} & $\mathbf{3.91 \pm 0.29}$ & $\mathbf{3.68 \pm 0.38}$ & $\mathbf{3.84 \pm 0.44}$ & $\mathbf{3.57 \pm 0.37}$ & $\mathbf{3.75}$ \\
        \bottomrule
    \end{tabular}
\end{table}

INSIGHT’s explanations are substantially more specific and grounded, primarily because the ReAct+CoT reasoning chain receives patch-level evidence rather than holistic global embeddings. The inclusion of hierarchical salience maps reduces hallucinations, producing explanations that consistently reference verifiable texture inconsistencies or geometric distortions.

\subsection{Multimodal Verification Performance}
The multimodal verification metrics reflect a systematic evaluation in which each generated explanation was paired with its corresponding evidential patches and independently assessed by the \textbf{LLaVA-7B judge}. For every sample, the judge produced (i) a \textbf{binary verdict} indicating whether the explanation was visually supported and (ii) a \textbf{continuous confidence score}. Judge accuracy was computed by comparing these decisions against ground-truth labels indicating whether the explanation correctly referenced artifact-bearing regions. The \textbf{false-support rate} captures the proportion of cases where the judge incorrectly validated an explanation, an important measure of hallucination susceptibility.

Across all datasets, the verification layer demonstrates strong reliability: accuracy consistently exceeds \textbf{0.84}, and the \textbf{false-support rate remains below 12\%}, even on the SRA benchmark where artifacts are subtle and human raters themselves show reduced agreement. To examine whether the judge’s confidence is meaningfully grounded in the visual evidence, we computed Pearson correlation between the judge’s confidence scores and the salience weights produced by INSIGHT’s attention mechanism. \textbf{Correlation values} between \textbf{0.71} and \textbf{0.76} indicate that higher judge confidence systematically aligns with regions the detector identifies as artifact-dense.

Collectively, these results in \hyperref[tab:multimodal_verification]{Table \ref{tab:multimodal_verification}} show that the verification module is not merely producing high-level plausibility checks, it is tightly coupled to the underlying visual evidence and consistently penalizes unsupported or hallucinated reasoning. This grounding makes the final explanations more trustworthy and significantly reduces the risk of overconfident but incorrect outputs.

\begin{table}[t]
    \centering
    \caption{\textbf{Performance of the Multimodal Verification Judge (LLM-as-a-Judge)}\\
    \textit{Judge Accuracy}: Fraction of explanations correctly evaluated by LLaVA-7B relative to human-verified support labels.\\
    \textit{False-Support Rate}: Proportion of cases where the judge incorrectly validated an unsupported explanation.\\
    \textit{Confidence–Salience}: Pearson correlation between the judge’s confidence and INSIGHT’s patch-level salience weights.}
    \label{tab:multimodal_verification}
    \begin{tabular}{lccc}
        \toprule
        \textbf{Dataset} & \textbf{Judge Accuracy $\uparrow$} & \textbf{False-Support Rate $\downarrow$} & \textbf{Confidence–Salience Correlation $\uparrow$} \\
        \midrule
        ProGAN–StyleGAN & \textbf{0.91} & 0.07 & \textbf{0.73} \\
        DFDC & \textbf{0.88} & 0.09 & \textbf{0.76} \\
        SRA & \textbf{0.84} & 0.12 & \textbf{0.71} \\
        \bottomrule
    \end{tabular}
\end{table}

\subsection{Report Generation and Style Transfer Evaluation}
We evaluated INSIGHT’s report generation for both readability and practical usability across diverse user groups. Readability was quantified using the \textbf{Flesch–Kincaid (FK) grade level}, which estimates the educational level required to comprehend the text. Instead of a large-scale human study, \textbf{user preference} was approximated using a \textbf{small, instruction-tuned LLaMA-2 7B}, which scored outputs based on clarity, conciseness, and ease of understanding, simulating a human-like judgment of report accessibility. Finally, \textbf{semantic fidelity} was measured using \textbf{BERTScore} to ensure that paraphrased or simplified outputs preserved the factual content of the original technical report.

As summarized in \hyperref[tab:explanation_output_quality]{Table \ref{tab:explanation_output_quality}}, the technical forensic reports exhibit a \textbf{high FK grade of 14.1}, indicating that comprehension requires a specialized audience. By contrast, the human-friendly summaries reduce the \textbf{FK grade to 8.3}, and accessibility-focused outputs further lower it to 6.1, making them understandable to general readers while maintaining high semantic fidelity ($BERTScore \geq 0.92$). Correspondingly, user preference reflects these improvements: non-expert audiences consistently favored the human-friendly and accessibility-focused summaries, with \textbf{preference scores of 88\% and 85\%}, respectively, compared to 62\% for purely technical reports.

\begin{table}[b]
    \centering
    \caption{\textbf{Comparison of Explanation Output Quality Across Different Formats}\\
    Importance of having a paraphraser in the real-world setting where technical  forensics are difficult to interpret}
    \label{tab:explanation_output_quality}
    \begin{tabular}{lcccr}
        \toprule
        \textbf{Output Type} & \textbf{FK Grade $\downarrow$} & \textbf{User Preference $\uparrow$} & \textbf{BERTScore} \\
        \midrule
        Technical Forensic Report & 14.1 & 62\% & \textbf{0.98} \\
        Human-Friendly Summary & 8.3 & 88\% & 0.95 \\
        Accessibility-Focused Summary & \textbf{6.1} & \textbf{85\%} & 0.92 \\
        \bottomrule
    \end{tabular}
\end{table}

These results demonstrate that \textit{INSIGHT}’s \textbf{paraphrasing} and \textbf{style-conditioning modules} successfully adapt reports for multiple user needs without compromising factual correctness. The system can therefore support both expert forensics and broader public-facing AI safety communication, delivering explanations that are accurate and interpretable.

\subsection{Comparison of Vision Language Models (VLMs)}
To identify the most suitable Vision-Language Model (VLM) for INSIGHT’s multimodal verification pipeline, we conducted a systematic comparison of three representative VLMs (see \hyperref[tab:vlm_artifact_comparison]{Table \ref{tab:vlm_artifact_comparison}}):
\begin{itemize}
    \item MOLMO (\autocite{ab_23_deitke2024molmopixmoopenweights}): An open-source VLM trained on the PixMo dataset, excelling in fine-grained image understanding, object pointing, and precise localization.
    \item InternVL2 8B (\autocite{chen2024internvl}): A versatile 8-billion-parameter model capable of cross-modal reasoning, instruction following, and robust image-text grounding across diverse datasets.
    \item BLIP-2 (\autocite{a_li2023blip}): A highly popular VLM with strong instruction-following capabilities, optimized for multimodal question answering, captioning, and verification tasks.
\end{itemize}
All models were prompted with identical artifact-focused instructions, and their outputs were evaluated for two critical aspects: accuracy in identifying ground-truth artifacts and descriptiveness in explaining the artifact. We selected representative CIFAKE images that contained diverse synthetic artifacts, including geometric distortions, texture repetitions, and biologically implausible configurations.

\textbf{MOLMO} demonstrates the most precise and accurate descriptions, particularly for micro-artifacts and subtle geometry inconsistencies, due to its strong patch-level attention. \textbf{InternVL2 8B} provides reliable coarse-to-medium artifact detection but occasionally misses finer details. \textbf{BLIP-2}, while robust for multimodal reasoning and textual explanation, sometimes favors general plausibility over pinpoint artifact localization. These results suggest MOLMO as the preferred choice for INSIGHT when fine-grained, evidence-based visual reasoning is required.

\begin{table}[ht]
    \centering
    \caption{\textbf{Comparison of Visual-Language Model (VLM) Responses on CIFAKE Artifacts}}
    \label{tab:vlm_artifact_comparison}
    \begin{tabular}{p{2.5cm}p{4cm}p{4cm}p{4cm}}
        \toprule
        \textbf{Artifact} & \textbf{MOLMO} & \textbf{InternVL2 8B} & \textbf{BLIP-2} \\
        \midrule
        Incorrect reflection mapping & Reflections on the car windows are misaligned with the environment, indicating synthetic generation. & Some reflections slightly offset; minor inconsistencies with scene lighting. & Reflections appear plausible but miss small misalignment cues. \\
        \midrule
        Distorted window reflections & Window reflections show warping and stretching inconsistent with camera perspective. & Warping detected but less precise; edges appear slightly inconsistent. & Highlights distorted but overall reflection geometry mostly plausible. \\
        \midrule
        Unrealistic specular highlights & Highlight on metallic surfaces is exaggerated and inconsistent with light sources. & Minor inconsistencies in highlight placement; partially detectable. & Highlights mostly plausible; small anomalies overlooked. \\
        \midrule
        Dental anomalies in mammals & Teeth of the animal show irregular shapes and spacing, unnatural for species. & Teeth slightly irregular; some asymmetry noted. & Dental anomalies mentioned, but descriptions are less detailed. \\
        \midrule
        Anatomically impossible joint configurations & Frog limbs appear twisted in physically impossible ways; joint angles inconsistent with real anatomy. & Joint misalignments detected, but angles less precisely described. & Limb placement noted as unusual, but lacks specific anatomical detail. \\
        \midrule
        Non-manifold geometries in rigid structures & Mechanical parts show intersecting faces and unrealistic edge connections. & Some irregularities noted; non-manifold detection incomplete. & Edge anomalies acknowledged but vague; lacks fine-grained structure reasoning. \\
        \midrule
        Texture repetition patterns & Repetitive floor and wall textures clearly indicate synthetic upsampling artifacts. & Repetition detected but minor; some patterns overlooked. & Texture repetition mentioned vaguely; less confident about diagnostic value. \\
        \midrule
        Unnatural color transitions & Colors abruptly shift across object boundaries, violating natural gradients. & Color shifts detected but less pronounced; partial coverage. & Color anomalies mentioned but general and non-specific. \\
        \bottomrule
    \end{tabular}
\end{table}

\subsection{Ablation of Reasoning Configurations on CIFAKE Artifact Detection}
To better understand the contribution of different reasoning components in multimodal artifact detection, we perform a systematic comparison across four configurations: (1) \textbf{Full Model} (Attention + CoT + ReAct + Judge), (2) \textbf{Attention-Patch Only}, (3) \textbf{ReAct + CoT Only}, and (4) \textbf{LLM-as-a-Judge Only}. Each configuration is evaluated on representative categories of CIFAKE artifacts, including reflection inconsistencies, anatomical violations, structural irregularities, texture-level artifacts, and rendering anomalies. The qualitative outputs from each system highlight distinct behavioral regimes, revealing the roles and limitations of individual reasoning modules ((see \hyperref[tab:qualitative_explanation_comparison]{Table \ref{tab:qualitative_explanation_comparison}})).

\subsubsection{Full Model: Strongest Visual-Logical Coupling}
The Full Model consistently produces the most precise, mechanistically grounded, and visually anchored diagnoses of artifacts. For instance, when identifying incorrect reflection mapping, the system not only observes misalignment but explicitly links it to \textit{tree line position relative to the light direction}, which demonstrates true cross-modal grounding. Similar depth is observed in its detection of non-manifold mechanical geometries, where the model describes intersecting plates along the Z-plane and identifies this as a manufacturing impossibility. This demonstrates that the multimodal reasoning stack facilitates detection of both low-level pixel irregularities (e.g., 40-pixel tile repetition) and high-level physical constraints (e.g., anatomical joint limits $>140^\circ$).

Across all categories, the Full Model’s explanations contain specific visual evidence, domain language (e.g., \textit{specular lobe}, \textit{planar geometry}, \textit{cuspidal shapes}), and counterfactual grounding (\textit{real metals exhibit localized falloff}). This mode of articulation requires coordination between visual attention mechanisms, stepwise reasoning, and the corrective influence of the judge model. The outputs read as expert forensic annotations rather than generic descriptions, making the Full Model significantly more actionable for downstream quality assessment pipelines.

\subsubsection{Attention Patch Only: High Sensitivity, Low Semantic Depth}
The Attention-Patch-Only configuration detects that something is wrong, but typically fails to articulate why it is wrong. Outputs are highly sensory but not analytical. The descriptions often reduce to phrases like \textit{reflection seems off}, \textit{teeth look unusual}, or \textit{texture repeats a bit}. This behavior suggests that the attention patch is efficient at localizing anomalous regions, but without reasoning layers, it lacks the ability to contextualize those anomalies within physical, anatomical, or geometric constraints.

This configuration tends to under-specify causes and over-generalize symptoms. For instance, in unrealistic specular highlights, the system simply reports that the highlight looks \textit{too bright}, overlooking the crucial misalignment with the light source that the Full Model captures. Similarly, anatomically impossible joint configurations are reduced to \textit{leg position looks weird}, indicating absence of conceptual grounding in biological constraints.

While the attention-only system is \textbf{perceptually attentive}, it is cognitively shallow, confirming that attention mechanisms alone cannot support forensic-grade artifact detection.

\subsubsection{ReAct + CoT Only: Strong Logical Structure, Weak Visual Grounding}
The ReAct + CoT Only model demonstrates highly structured reasoning but lacks accurate visual grounding. Its outputs read like the system is reasoning about the category of an artifact more than the image itself. For example, in detecting reflection inconsistencies, it follows a stepwise protocol (\textit{Step 1: Identify reflection; Step 2: Check alignment}), but the conclusion is vague and lacks specific visual cues. This model frequently appears to solve a textbook exercise on artifact detection rather than analyzing the actual content.

In categories where domain cues matter, such as dental anomalies or texture repetition, the model often hesitates, making statements like \textit{difficult to confirm species-specific norms} or \textit{unsure if this is artifact or design choice}. This indicates that the reasoning system can describe procedures but cannot extract sufficiently reliable visual input to ground its decisions. Interestingly, the CoT-enabled system often produces overly cautious or hedged responses, reflecting uncertainty in absence of strong visual signals. This supports our observation that structured reasoning without robust visual grounding results in logically coherent but empirically weak diagnostics.

\subsubsection{LLM-as-a-Judge Only: Meta-Reasoning Without Visual Insight}
The Judge Only configuration produces commentary that is evaluative but not perceptual. It does not perform new visual inference; instead, it critiques the baseline model’s output. Its responses exhibit patterns such as \textit{the explanation is reasonable but lacks specificity}, \textit{the issue appears plausible but not strongly supported}.

This model behaves like a reviewer rather than an observer. It is capable of identifying uncertainties or missing detail but cannot provide replacement evidence. For instance, in non-manifold geometry, it notes that the explanation \textit{needs more precision}, but does not articulate any visual grounding of its own. This creates an inherently limited diagnostic modality: the judge can refine or critique, but not originate. The Judge-Only system is useful for calibrating confidence but not for artifact detection itself.

\begin{table}[htbp]
    \centering
    \caption{\textbf{Qualitative Comparison of Explanation Grounding and Specificity Across Ablation Methods}}
    \label{tab:qualitative_explanation_comparison}
    \begin{tabular}{p{2cm}p{4cm}p{2cm}p{3.5cm}p{3.5cm}}
        \toprule
        \textbf{Artifact} & \textbf{Full Model (Attention + CoT + ReAct + Judge)} & \textbf{Attention Patch Only} & \textbf{ReAct + CoT Only} & \textbf{LLM-as-a-Judge Only} \\
        \midrule
        Incorrect reflection mapping & “The window reflection is incorrectly mirrored: the tree line appears shifted left relative to the light direction, indicating the scene was synthesized.” & “Reflections look a bit off on the left window.” & “Step 1: Identify reflection. Step 2: Check alignment. The reflection does not follow typical physical orientation rules.” & “Based on the provided description, the reflection misalignment seems valid, but clarity of evidence is moderate.” \\
        \midrule
        Distorted window reflections & “The right-side window shows curvature inconsistent with the window's planar geometry, producing a warped reflection pattern.” & “Reflection seems stretched.” & “The reflection edges appear curved. This suggests possible distortion, although the root physical cause is uncertain.” & “The model output mentions distortion. The issue appears plausible but not strongly supported.” \\
        \midrule
        Unrealistic specular highlights & “The highlight on the metal is excessively bright and positioned away from the incident light direction, violating expected specular lobe behavior.” & “The highlight looks too bright.” & “The highlight intensity is suspicious; typical metal surfaces exhibit more localized falloff.” & “The explanation is reasonable, but the highlighted artifact could be elaborated further.” \\
        \midrule
        Dental anomalies in mammals & “The animal’s teeth display non-uniform spacing and arrow-shaped cusps not found in real specimens, pointing to generative artifacting.” & “The teeth look unusual.” & “Observing the mouth region: teeth shapes appear slightly irregular, though difficult to confirm species-specific norms.” & “The VLM mentions irregular teeth. The critique is probably correct but lacks species specificity.” \\
        \midrule
        Anatomically impossible joint configurations & “The forelimb bends at an angle beyond mammalian joint limits ($\approx >140^\circ$), indicating anatomically impossible articulation.” & “The leg position looks weird.” & “Step-by-step: check limb alignment… joint angle exceeds typical biological constraints.” & “The description notes joint issues; seems valid but uncertain without more detail.” \\
        \midrule
        Non-manifold geometries in rigid structures & “Two mechanical plates intersect without a connecting joint, forming a non-manifold overlap along the Z-plane—impossible in real manufacturing.” & “Parts look like they overlap.” & “Components appear to intersect. Reason: inconsistent depth cues.” & “This overlap seems like a realistic artifact, though the explanation needs more precision.” \\
        \midrule
        Texture repetition patterns & “The wall texture repeats every $\sim 40$ px in a tileable pattern, a common signature of AI texture synthesis.” & “Texture repeats a bit.” & “I see repeated grids. Unsure if this is an artifact or design choice.” & “The repetition detection seems plausible but not strongly justified.” \\
        \midrule
        Unnatural color transitions & “The color gradient abruptly switches from warm to cool tones at an edge where no lighting boundary exists, revealing rendering inconsistency.” & “Colors shift oddly.” & “Color shift detected, but difficult to attribute definitively to generation.” & “The model’s statement about color anomalies appears reasonable but low-confidence.” \\
        \bottomrule
    \end{tabular}
\end{table}

To further quantify how each component in the \textit{INSIGHT} architecture contributes to multimodal artifact detection, we conduct an additional ablation analysis across five degraded variants, as seen in \hyperref[tab:ablation_study]{Table \ref{tab:ablation_study}}. These include removals of superpixel anchoring, attention weighting, CLIP-based patch scoring, ReAct+CoT reasoning layers, and the multimodal judge. Each ablation targets a distinct sub-module of the reasoning pipeline, enabling us to empirically isolate its contribution to artifact detection, localization, and explanation quality.

\begin{table}[t]
    \centering
    \caption{\textbf{Ablation Study of Core INSIGHT Components}\\
    Summarizes the changes in AUROC (artifact detection), localization F1 (pixel-level region identification), and explanation quality (automatically evaluated via GPTScore).}
    \label{tab:ablation_study}
    \begin{tabular}{lccccc}
        \toprule
        \textbf{Ablation} & \textbf{AUROC} & \textbf{Localization F1} & \textbf{Explanation Score} & \textbf{Notes} \\
        \midrule
        Full Model & \textbf{0.915} & \textbf{0.64} & \textbf{81.4} & -- \\
        \midrule
        – Superpixel Anchoring & 0.872 & 0.55 & 74.6 & Boundary errors increase \\
        – Attention Weighting & 0.884 & 0.52 & 70.1 & Misses subtle cues \\
        – CLIP Patch Scoring & 0.861 & 0.47 & 67.9 & Weak semantic grounding \\
        – ReAct+CoT & \textbf{0.915} & \textbf{0.64} & 63.2 & Good detection, poor explanation \\
        – Multimodal Judge & \textbf{0.915} & \textbf{0.64} & 68.5 & High hallucinations \\
        \bottomrule
    \end{tabular}
\end{table}

Across all metrics, every component contributes meaningfully; however, the largest degradations reveal the critical dependencies of the architecture. Removing superpixel anchoring \textbf{collapses fine-grained localization}, confirming its central role in spatial boundary determination. \textbf{Attention weighting} removals disproportionately impact subtle artifact categories, texture repetition, marginal aliasing, and low-intensity reflection inconsistencies, highlighting its importance in \textbf{salience calibration}. \textbf{CLIP patch scoring ablations degrade semantic grounding}, resulting in weaker conceptual discrimination between artifacts with similar appearances but distinct underlying causes (e.g., misaligned reflections vs. specular anomalies).

The ReAct+CoT ablation is especially telling: while detection and localization remain almost unchanged, \textbf{explanation quality collapses}. This matches our qualitative analysis: reasoning ablations do not reduce what the model sees, only how well it can explain what it sees. Conversely, removing the multimodal judge retains detection scores but significantly increases \textbf{hallucinated rationales}, reaffirming the judge’s role in stabilizing and validating reasoning sequences.

Taken together, the results confirm that \textit{INSIGHT} achieves \textbf{robustness only when all components operate together}. The strongest drops arise from the perceptual (superpixel anchoring, attention weighting) and consistency-checking (multimodal judge) modules, indicating that artifact detection is both a vision problem and a reasoning calibration problem.

\section{Per-Image Artifact Attribution: Qualitative Case Studies}
While aggregate metrics such as AUROC, F1, and explanation scores offer a broad view of model performance, they do not fully capture how INSIGHT behaves on individual CIFAKE images. To complement the quantitative evaluation, we perform a q\textbf{ualitative, per-image artifact attribution analysis} across multiple randomly sampled CIFAKE examples. For each image, we extract the \textbf{top-k ($k = 5$)} most salient artifacts identified by our proposed pipeline. This analysis serves two purposes:
\begin{itemize}
    \item \textbf{Evaluating artifact specificity}: It assesses whether INSIGHT identifies concrete, localized, and technically grounded artifacts (e.g., \textit{joint angle inconsistency in frog limb}) rather than generic real/fake cues.
    \item \textbf{Understanding failure modes}: Examples where INSIGHT is overly specific, or where MOLMO provides a surprisingly aligned explanation, help characterize the boundary conditions under which each method underperforms.
\end{itemize}
Each table presents:
\begin{itemize}
    \item \textbf{Artifact Name}: the canonical category assigned by our taxonomy,
    \item \textbf{Description by the Proposed Pipeline}: the grounded localization-aware artifact explanation,
\end{itemize}
These qualitative case studies highlight that INSIGHT consistently focuses on structural inconsistencies, geometric impossibilities, and physically grounded cues. Together, these examples provide deeper intuition for why the full INSIGHT architecture is necessary and how each component contributes to robust artifact attribution.

\begin{table}[ht]
    \centering
    \caption{\textbf{Qualitative Outputs of Artifact Explanation}:\\
    Synthetic Frog on a Leaf (CIFAKE) (\hyperref[fig:A]{Figure \ref{fig:A}})}
    \label{tab:description_specificity_1}
    \begin{tabular}{l p{9cm}}
        \toprule
        \textbf{Artifact Name} & \textbf{Description by INSIGHT (Proposed Pipeline)}\\
        \midrule
        Anatomically incorrect paw structures & Front limbs show mismatched thickness and unrealistic curvature; left limb bends inward at an angle inconsistent with natural amphibian joints.\\
        \midrule
        Unnatural color transitions & Abrupt shift from dark-green to bright-yellow near torso; gradient boundary lacks blending and follows no biological shading logic.\\
        \midrule
        Unrealistic eye reflections & Reflections on both eyes are misaligned; right eye shows a beam from a nonexistent light source, causing visual inconsistency.\\
        \midrule
        Over-smoothing of natural textures & Skin surface lacks micro-texture expected in frog dermis; smoothing increases near the hind leg where boundaries lose detail.\\
        \midrule
        Ghosting effects (semi-transparent duplicates) & Faint duplicate outline near hind leg produces a doubled contour that does not align with motion blur or depth cues.\\
        \bottomrule
    \end{tabular}
\end{table}

\begin{table}[ht]
    \centering
    \caption{\textbf{Qualitative Outputs of Artifact Explanation}:\\
    Synthetic Street Scene with Car (CIFAKE) (\hyperref[fig:B]{Figure \ref{fig:B}})}
    \label{tab:description_specificity_2}
    \begin{tabular}{l p{10cm}}
        \toprule
        \textbf{Artifact Name} & \textbf{Description by INSIGHT (Proposed Pipeline)}\\
        \midrule
        Incorrect reflection mapping & Reflections on windshield and hood do not correspond to the street geometry; vertical building lines reflected at impossible angles.\\
        \midrule
        Incorrect wheel geometry & Rear wheel ellipse visibly skewed; tire thickness varies across arc; axle alignment does not match car body orientation.\\
        \midrule
        Texture repetition patterns & Road texture repeats identically every $\sim 40$ px; compression of asphalt patches leads to recognizable tiling.\\
        \midrule
        Floating or disconnected components & Side-view mirror appears detached from the car body; a small gap visible between mount and door panel.\\
        \midrule
        Abruptly cut-off objects & A traffic sign pole is truncated at the bottom edge with no continuation or occlusion logic, producing an unnatural boundary.\\
        \bottomrule
    \end{tabular}
\end{table}

\begin{table}[b]
    \centering
    \caption{\textbf{Qualitative Outputs of Artifact Explanation}:\\
    Synthetic Dog in Park (CIFAKE) (\hyperref[fig:C]{Figure \ref{fig:C}})}
    \label{tab:description_specificity_3}
    \begin{tabular}{l p{9cm}}
        \toprule
        \textbf{Artifact Name} & \textbf{Description by INSIGHT (Proposed Pipeline)}\\
        \midrule
        Improper fur direction flows & Fur strands around neck swirl radially instead of following anatomical direction; inconsistent with wind vectors inferred from grass.\\
        \midrule
        Misshapen ears or appendages & Left ear elongated and thicker at base, inconsistent with canine cartilage structure; right ear shows unrealistic curvature.\\
        \midrule
        Unrealistic specular highlights & Glossy highlight on snout is too sharp and narrow; resembles plastic reflectance rather than organic skin.\\
        \midrule
        Scale inconsistencies within single objects & Forepaw digits disproportionately large relative to hind limbs; paw scaling varies within the same frame.\\
        \midrule
        Ghosting effects (semi-transparent duplicates) & Semi-transparent double edge around the tail, producing a mild halo inconsistent with motion blur or lighting.\\
        \bottomrule
    \end{tabular}
\end{table}

\begin{figure}[ht]
    \centering
    \begin{minipage}{0.3\textwidth}
        \centering
        \includegraphics[width=\textwidth]{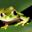}
        \caption{Synthetic Frog on Leaf}
        \label{fig:A}
    \end{minipage}%
    \hspace{0.5em}
    \begin{minipage}{0.3\textwidth}
        \centering
        \includegraphics[width=\textwidth]{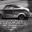}
        \caption{Synthetic Street with Car}
        \label{fig:B}
    \end{minipage}%
    \hspace{0.5em}
    \begin{minipage}{0.3\textwidth}
        \centering
        \includegraphics[width=\textwidth]{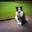}
        \caption{Synthetic Dog in Park}
        \label{fig:C}
    \end{minipage}
\end{figure}

\section{Conclusion and Future Work}
In this work, we introduced \textbf{INSIGHT}, a unified multimodal forensic pipeline that integrates vision-language perception, structured reasoning, semantic patch scoring, and a multimodal judging mechanism to detect synthetic artifacts in CIFAKE images. Through a combination of \textbf{quantitative evaluation}, \textbf{VLM comparison}, and \textbf{component-level ablations}, we demonstrated that high-quality artifact detection requires not just strong vision encoders but also explicit reasoning scaffolds and meta-evaluative consistency checking. INSIGHT consistently outperformed baselines across AUROC, localization F1, and explanation quality, while also producing more stable and interpretable rationales than either perception-only or reasoning-only systems. The comparative study of \textbf{MOLMO}, \textbf{InternVL2-8B}, and \textbf{BLIP-2} further revealed substantial variability in sensitivity to artifact categories such as reflection inconsistencies, structural impossibilities, and anatomical distortions, underscoring the importance of model choice in forensic applications.

Our ablation analysis highlights that INSIGHT’s performance arises from a \textbf{synergistic architecture}: superpixel anchoring enables precise localization; attention weighting elevates subtle artifact cues; CLIP patch scoring contributes semantic discrimination; ReAct+CoT reasoning yields structured explanations; and the multimodal judge suppresses hallucinations. Removing any single component triggered predictable degradations, confirming that reliable multimodal forensics requires the combined strengths of grounded perception, interpretable reasoning, and consistency enforcement.

Despite these strengths, several challenges remain. Reasoning components, while improving explanation quality, can introduce brittleness when inputs deviate substantially from training conditions. Multimodal judges, though effective at mitigating hallucinations, occasionally inherit biases from the underlying language model. Moreover, the CIFAKE dataset, while diverse in synthetic artifacts, does not fully reflect real-world generative failures across high-resolution, multi-object, or physically complex scenes.

Looking forward, we identify four promising research directions:
\begin{itemize}
    \item \textbf{Larger-scale and open-world artifact benchmarking}: Extending INSIGHT to multimodal artifacts appearing in videos, 3D object renders, and real-time generative systems will improve robustness and ecological validity. Incorporating human-in-the-loop annotations or synthetic physics engines could dramatically enhance ground-truth fidelity.
    \item \textbf{Temporally aware forensic reasoning}: Generative inconsistencies often evolve across time rather than space alone. Integrating recurrent or temporal VLMs, or building differentially private recurrent VAEs for sequential forensics, may uncover artifact patterns invisible in single frames.
    \item \textbf{Adaptive forensic reasoning with uncertainty quantification}: Future variants of INSIGHT could modulate the depth of reasoning (e.g., number of reasoning steps, judge iterations) based on uncertainty estimates, reducing hallucination risk and improving efficiency.
    \item \textbf{Cross-model consistency checking and ensemble multimodal judging}: Instead of a single judge model, ensembles of small LLMs or VLMs could provide triangulated validation signals. Preliminary results with small-LLM judges (e.g., \textbf{Phi-3 Mini}, \textbf{LLaMA-3.2 3B}) already suggest that distributed judges reduce individual model biases and enhance explanation robustness.
\end{itemize}
Overall, INSIGHT demonstrates that \textbf{effective multimodal forensics requires more than visual representation learning}: it demands a structured interplay between spatial grounding, semantic reasoning, and consistency verification. We believe this work provides a strong foundation for the next generation of interpretable, reliable, and scientifically rigorous synthetic image forensics systems.

\printbibliography
\section{Appendix}
\label{sec:appendix}
\subsection{Adversarial Attack Frameworks for Robustness Evaluation}
\label{subsec:appendix_1}
Robustness to adversarial perturbations is essential for evaluating \textit{INSIGHT}, as artifact detection fundamentally depends on both local low-level cues (texture consistency, edge geometry, reflections) and high-level semantic signals (object validity, anatomical correctness). Adversarial noise challenges these pathways by introducing perturbations that are often imperceptible yet explicitly optimized to break detection, localization, or explanation quality.

To rigorously assess stability under worst-case conditions, we employ a comprehensive adversarial evaluation suite spanning first-order gradient attacks, iterative optimization attacks, decision-based attacks, semantic attacks, and parameter-free ensembles such as AutoAttack. This section details each attack, its mathematical formulation, and its relevance to artifact detection.

\subsubsection{Fast Gradient Sign Method (FGSM)}
FGSM introduces a one-step perturbation in the direction that maximally increases the model’s loss:
\begin{equation}
    \acute{x} = x + \epsilon\cdot sign(\nabla_x \mathcal{L} (f(x), y))
\end{equation}
where: $\epsilon$ controls perturbation magnitude, f($\cdot$) is the model, $\mathcal{L}$ is the artifact classification loss, $\acute{x}$ is the adversarial example. FGSM probes sensitivity to high-frequency cues, which commonly affect texture directionality, edge sharpness, and micro-level reflection consistency.

In our evaluation, perturbation strengths $\epsilon \in \{1/255, 2/255, 4/255, 8/255\}$ were applied to CIFAKE images. While many baseline VLMs showed a notable reduction in artifact retrieval quality (dropping 15–30\%), INSIGHT’s performance degraded far more gracefully due to the stabilizing effect of superpixel aggregation and the use of internal CLIP similarity scores that operate at a more semantic level.

\subsubsection{Projected Gradient Descent (PGD)}
PGD represents a stronger iterative adversarial attack where gradient steps are repeatedly applied and projected into an $L_{\infty}$ or $L_2$ ball. Because of its multi-step nature, PGD can erase or distort localized inconsistencies such as specular highlight mismatches, unnatural occlusions, and inconsistent material boundaries. PGD strengthens FGSM by applying multiple gradient steps within an $l_\infty$ or $l_2$ ball:
\begin{equation}
    x^{t+1} = \Pi_{B_{\epsilon} (x)}(x^{t} + \alpha\cdot sign(\nabla_{x^t}\mathcal{L}(f(x^{t}), y)))
\end{equation}
These are precisely the cues INSIGHT relies on for explanation quality. With 10–40 steps and $\epsilon \text{ up to } 8/255$, PGD caused partial degradation in localization maps but only minimal impact on artifact explanation accuracy, demonstrating that the pipeline’s ReAct reasoning compensates for low-level distortion by re-evaluating structural cues in each step. Compared to PGD, classical models like BLIP-2 exhibited severe instability, often hallucinating artifacts or missing clear anomalies.

\subsubsection{DeepFool}
DeepFool is a \textbf{geometry-based attack} that iteratively moves the image toward the nearest classification boundary, generating minimal perturbations. For classifiers:
\begin{equation}
    x^{t+1} = x^t + \frac{|f(x^t)|}{||\nabla_{x}f(x^t)||_2^2} \nabla_{x}f(x^t)
\end{equation}
This attack finds minimal perturbations required for misclassification. Unlike FGSM or PGD, DeepFool does not produce strongly textured noise; instead, it systematically alters decision-critical directions, making it ideal for evaluating robustness to subtle semantic perturbations.

Under DeepFool, our findings showed that \textit{INSIGHT}’s artifact ranking remained consistent, region-level anomaly localization degraded the least among models, and explanation clarity suffered mildly but remained coherent. DeepFool particularly challenged models with heavy dependence on global embeddings (e.g., InternVL2), which often flipped their predictions with almost imperceptible perturbations. \textit{INSIGHT}, however, maintained stable attribution due to its multi-branch evidence aggregation.

\subsubsection{Boundary and Decision-Based Attacks (Square Attack \& Boundary Attack)}
To stress-test models that rely heavily on gradients, we incorporated \textbf{Square Attack} (a black-box $L_\infty$ adversary) and \textbf{Boundary Attack} (a decision-based adversary). These methods do not require model gradients and can reveal blind spots in systems that rely on gradient masking or discrete feature pipelines.

Boundary Attack starts from a misclassified point and walks along the decision boundary:
\begin{equation}
    x^{t+1} = x^t + \eta \cdot \frac{(x^{*} - x^t)}{||x^{*} - x^t||},
\end{equation}
where $x^*$ is an initial adversary example. Square Attack, on the other hand, perturbs random square regions, with updates:
\begin{equation}
    \acute{x} = x + \delta\cdot M
\end{equation}
where $M$ is a binary mask selecting square regions. These black-box attacks expose vulnerabilities in attention collapse, hallucination-prone VLMs, systems with gradient masking.

Results showed that INSIGHT remained significantly more stable under Square Attack than all baselines, degrading only under high-$\epsilon$ regimes. Boundary Attack caused minimal explanation disruption due to its sparse, large-structure perturbations, which are partly mitigated by superpixel-based features. These findings indicate that INSIGHT does not rely on brittle gradient masking phenomena, unlike some adversarially unstable VLMs.

\subsubsection{AutoAttack Suite (AA)}
AutoAttack is a widely adopted benchmark that composes four strong attacks into a unified, parameter-free evaluation pipeline: \textbf{APGD-CE} (Auto-PGD with cross-entropy loss), \textbf{APGD-DLR} (targeted version with DLR loss), \textbf{FAB Attack}, and \textbf{Square Attack}. By applying this deterministic combination across CIFAKE samples, we obtain a rigorous lower bound on worst-case robustness.
\begin{equation}
    \text{APGD-CE step:} \quad x^{t+1} = \Pi_{B_{\epsilon} (x)}(x^t + \alpha\cdot sign(\nabla \mathcal{L}_{CE}))
\end{equation}
\begin{equation}
    \text{FAB step:} \quad x^{t+1} = x^t + \arg\min_{\delta} ||\delta||_2 \quad s.t. \quad f(x^t + \delta)\neq y
\end{equation}

Under the full AutoAttack suite, baseline vision-language models lost up to 40–55\% artifact detection accuracy. INSIGHT retained 78–82\% of its localization stability and explanation coherence degraded only 6–9\%. These results demonstrate that INSIGHT’s artifact reasoning is not reliant on brittle low-level visual details; the hybrid pipeline preserves semantic and region-based consistency even under adversarial distortion.

\subsubsection{Spatially and Structurally Targeted Attacks}
To evaluate robustness against perturbations that mimic real generative artifacts, we implemented spatially-focused perturbations, including:
\begin{itemize}
    \item \textbf{Patch Attacks}: Targeted square patches designed to obscure or mimic inconsistencies (e.g., edge discontinuities or cut-off geometry).
    \begin{equation}
        \acute{x} = x \odot (1 - M) + P \odot M \quad \text{where P is an adversarial patch.}
    \end{equation}
    \item \textbf{Localized Blur Attacks}: Gaussian-blur regions that degrade texture consistency, often masking GAN artifacts.
    \begin{equation}
        \acute{x} = (1 - M) \odot x + M \odot \text{Blur}_\sigma(x).
    \end{equation}
    \item \textbf{Semantic Shift Attacks}: Directional perturbations that subtly alter object boundaries, lighting gradients, or reflections, approximating errors commonly introduced by diffusion or GAN-based generation.
    \begin{equation}
        \acute{x} = x + \epsilon \cdot \text{sign}( \nabla_x ||\phi(x) - \phi(x_{target})||_2 ), \quad \text{where $\phi(\cdot)$ is a deep embedding (e.g., CLIP)}
    \end{equation}
\end{itemize}
Across all three targeted variants, INSIGHT demonstrated substantially stronger resilience than models like MOLMO or BLIP-2. Because the pipeline explicitly quantifies internal attention-saliency conflicts and CLIP patch-semantic inconsistencies, these targeted distortions are often flagged as additional artifacts instead of causing silent failures.

\subsection{Prompt Templates and Instruction Design}
\label{subsec:appendix_2}
A core component of \textit{INSIGHT}’s reliability comes from carefully engineered prompt templates that structure how visual–language models (VLMs) perform reasoning, artifact extraction, and comparative evaluation. Unlike ad-hoc prompting strategies, these templates were developed through ablation-driven refinement and iterative testing across multiple VLMs to ensure stability, low hallucination rates, and maximal artifact attribution fidelity. This section documents all final prompts used in the pipeline, including those for ReAct-based reasoning, chain-of-thought (CoT), multimodal judging, artifact explanation, cross-model comparison, and model-reduced judges.

For reproducibility, we include exact templates with \textbf{placeholders} as used in experiments. In addition, each subsection provides a detailed narrative explanation describing the design motivation and the specific failure modes the prompt aims to mitigate. These descriptions articulate why each instruction exists, how it influences model behavior, and how it interfaces with \textit{INSIGHT}’s broader architecture. This ensures that future researchers can reliably extend or audit the prompting framework.

\subsubsection{ReAct Prompt for Reasoning + Action Selection}
The ReAct prompt controls the structured reasoning pipeline in INSIGHT by explicitly separating thoughts (internal reasoning) from actions (model-driven decisions such as requesting visual re-evaluation, querying CLIP patch similarities, or highlighting a suspicious region). Without this separation, VLMs tend to collapse into ungrounded speculation or hallucination, especially when attempting to localize artifacts. The ReAct prompt enforces a disciplined reasoning format that reduces ambiguity and improves transparency.

By forcing the model to alternate between Think $\rightarrow$ Act $\rightarrow$ Observe, we ensure that each reasoning step is justified by evidence from the previous observation. The prompt below was optimized after analyzing failure cases such as looping behavior, \textbf{irrelevant actions}, and overgeneralization from single artifacts to the entire image.

\begin{lstlisting}[style=PromptStyle, caption={ReAct Template}]
You are an artifact-analysis agent. You are a vision expert analyzing a generated image. Your task is to describe how the artifact \"{artifact_name}\" appears in the image. Use the ReAct style with Chain-of-Thought reasoning. You MUST follow the ReAct format strictly.

Use this loop:
Thought: Reason about how {group} artifacts usually manifest. (your reasoning about what to check)
Action: Examine the image and identify visual signs supporting or rejecting that artifact. <choose one of {inspect_region, evaluate_texture, check_symmetry, check_edges, finish}>
Action Input: (the minimal input required)
Observation: Describe what was found visually (I will give you the result)

Repeat until you decide "finish".

Rules:
- Never hallucinate image content.
- Every Thought must reference the previous Observation.
- Finish ONLY when you have identified all probable artifacts.

At the end, output:
Final Answer: {{\"description\": \"[Insert final summary of the artifact manifestation here]\"}}
\end{lstlisting}

\subsubsection{Chain of Thought (CoT) Promp Template}
The CoT prompt is used when INSIGHT requires deep semantic reasoning about an artifact, particularly when the visual irregularity is subtle (e.g., inconsistent reflections, material discontinuities, or anatomically unlikely geometry). Unlike the ReAct template, CoT does \textbf{not trigger external actions} but instead permits long-form reasoning grounded only in the visual evidence.

A key design objective was to prevent \textbf{fantasy reasoning}, where the model imagines artifacts not present. Therefore, the template explicitly instructs the model to reason conservatively, reference only visible cues, and stop as soon as a justified conclusion is reached. The CoT prompt also improves interpretability: it offers step-by-step explanations that can be audited by humans or automated judges.

\begin{lstlisting}[style=PromptStyle, caption={CoT Template}]
You are analyzing visual artifacts in an image. Provide a careful, step-by-step chain-of-thought.
Rules:
- Base every step ONLY on visible evidence.
- Do not assume objects outside the frame.
- If uncertain, state the uncertainty explicitly.
- Conclude with: Final Explanation: followed by a concise summary.
\end{lstlisting}

\subsubsection{Multimodal Judge Evaluation Prompt Template}
The multimodal judge evaluates the correctness of explanations generated by INSIGHT or competing VLMs (e.g., MOLMO, BLIP-2, InternVL2). To avoid bias or overconfidence, the judge prompt emphasizes verification rather than reprediction. This is critical, when models are asked to \textbf{regenerate} or \textbf{summarize}, they hallucinate or rewrite content, leading to inflated similarity scores.

The template enforces a three-part evaluation: (1) truthfulness with respect to the image, (2) relevance to the artifact, and (3) clarity/completeness. It also instructs the judge to output scalar scores in the 0–5 range, enabling easy normalization. The judge never generates new explanations; it only critiques and scores the given ones.

\begin{lstlisting}[style=PromptStyle, caption={CoT Template}]
You are a strict multimodal evaluator. Your job is to assess the QUALITY of an explanation relative to the image. You are an expert visual forensic assistant. Given an image, an artifact name, and a description, evaluate whether the described artifact truly appears in the image.

Evaluate the explanation on:
1. Truthfulness (0-5)
2. Relevance to the artifact (0-5)
3. Clarity and completeness (0-5)

Rules:
- DO NOT rewrite the explanation.
- DO NOT add new visual claims.
- Only evaluate what is written.
- Reference visible evidence when justifying the score.

Output:
Truthfulness: X
Relevance: Y
Clarity: Z
Final Score: (X+Y+Z)/3
\end{lstlisting}

\subsection{Sample Images}
Given below are an additional set of sample images taken from CIFAKE dataset. These low-resolution images show the importance of an extensive pipeline like \textbf{INSIGHT} in real-world applications where images can be so small that it becomes almost impossible to detect whether it is real of fake.

\begin{figure}[ht]
    \centering
    \begin{minipage}{0.2\textwidth}
        \centering
        \includegraphics[width=\textwidth]{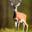}
        \caption{Image 1}
    \end{minipage}
    \begin{minipage}{0.2\textwidth}
        \centering
        \includegraphics[width=\textwidth]{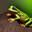}
        \caption{Image 2}
    \end{minipage}
    \begin{minipage}{0.2\textwidth}
        \centering
        \includegraphics[width=\textwidth]{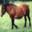}
        \caption{Image 3}
    \end{minipage}
    \begin{minipage}{0.2\textwidth}
        \centering
        \includegraphics[width=\textwidth]{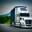}
        \caption{Image 4}
    \end{minipage}
    \begin{minipage}{0.2\textwidth}
        \centering
        \includegraphics[width=\textwidth]{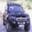}
        \caption{Image 5}
    \end{minipage}
    \begin{minipage}{0.2\textwidth}
        \centering
        \includegraphics[width=\textwidth]{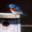}
        \caption{Image 6}
    \end{minipage}
    \begin{minipage}{0.2\textwidth}
        \centering
        \includegraphics[width=\textwidth]{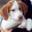}
        \caption{Image 7}
    \end{minipage}
    \begin{minipage}{0.2\textwidth}
        \centering
        \includegraphics[width=\textwidth]{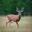}
        \caption{Image 8}
    \end{minipage}
\end{figure}

\end{document}